\documentclass[10pt,twocolumn,letterpaper]{article}

\usepackage{cvpr}              % To produce the CAMERA-READY version
\usepackage{multirow}
\usepackage{makecell}
\usepackage{xcolor}         % colors
\usepackage{gensymb}
\usepackage{tabularx}   % in preamble
\usepackage{booktabs}   % for \toprule etc.
\usepackage{array}      % for custom X column
\usepackage{pifont}

\newcolumntype{Y}{>{\raggedright\arraybackslash}X} % ragged-
\definecolor{cvprblue}{rgb}{0.21,0.49,0.74}
\usepackage[pagebackref,breaklinks,colorlinks,allcolors=cvprblue]{hyperref}

\def\paperID{30338} % *** Enter the Paper ID here
\def\confName{CVPR}
\def\confYear{2026}

\definecolor{cvprblue}{rgb}{0.21,0.49,0.74}
\usepackage[pagebackref,breaklinks,colorlinks,allcolors=cvprblue]{hyperref}

\def\paperID{30338} % *** Enter the Paper ID here
\def\confName{CVPR}
\def\confYear{2026}

\newcommand{\parvspace}{\vspace{-11pt}}

\title{GazeShift: Unsupervised Gaze Estimation and Dataset for VR}

\author{
Gil Shapira$^{1}$ \quad 
Ishay Goldin$^{1}$ \quad 
Evgeny Artyomov$^{1}$ \quad \\
Donghoon Kim$^{2}$ \quad 
Yosi Keller$^{3}$ \quad 
Niv Zehngut$^{1}$
\and
$^{1}$Samsung Semiconductor Israel R\&D Center (SIRC) \\
$^{2}$Samsung Electronics \quad $^{3}$Bar-Ilan University \\
{\tt\small \{gil.shapira, ishay.goldin, evgeny.a, dhoon.kim, niv.z\}@samsung.com} \\
{\tt\small yosi.keller@gmail.com}
}
\begin{document}
\maketitle

\begin{abstract}
Gaze estimation is instrumental in modern virtual reality (VR) systems. Despite significant progress in remote-camera gaze estimation, VR gaze research remains constrained by data scarcity, particularly the lack of large-scale, accurately labeled datasets captured with the off-axis camera configurations typical of modern headsets. Gaze annotation is difficult since fixation on intended targets cannot be guaranteed. To address these challenges, we introduce VRGaze, the first large-scale off-axis gaze estimation dataset for VR, comprising 2.1 million near-eye infrared images collected from 68 participants. We further propose GazeShift, an attention-guided unsupervised framework for learning gaze representations without labeled data. Unlike prior redirection-based methods that rely on multi-view or 3D geometry, GazeShift is tailored to near-eye imagery, achieving effective gaze-appearance disentanglement in a compact, real-time model. GazeShift embeddings can be optionally adapted to individual users via lightweight few-shot calibration, achieving a 1.84° mean error on VRGaze. On the remote-camera MPIIGaze dataset, the model achieves a 7.15° person-agnostic error, doing so with 10x fewer parameters and 35x fewer FLOPs than baseline methods. Deployed natively on a VR headset GPU, inference takes only 5 ms. Combined with demonstrated robustness to illumination changes, these results highlight GazeShift as a label-efficient, real-time solution for VR gaze tracking. Project code and the VRGaze dataset are released at \url{https://github.com/gazeshift3/gazeshift}

\end{abstract}
\vspace{-1em} 

\begin{figure*}[h]
\includegraphics[clip, trim=1cm 4.5cm 3.7cm 2cm,width=0.9\linewidth]{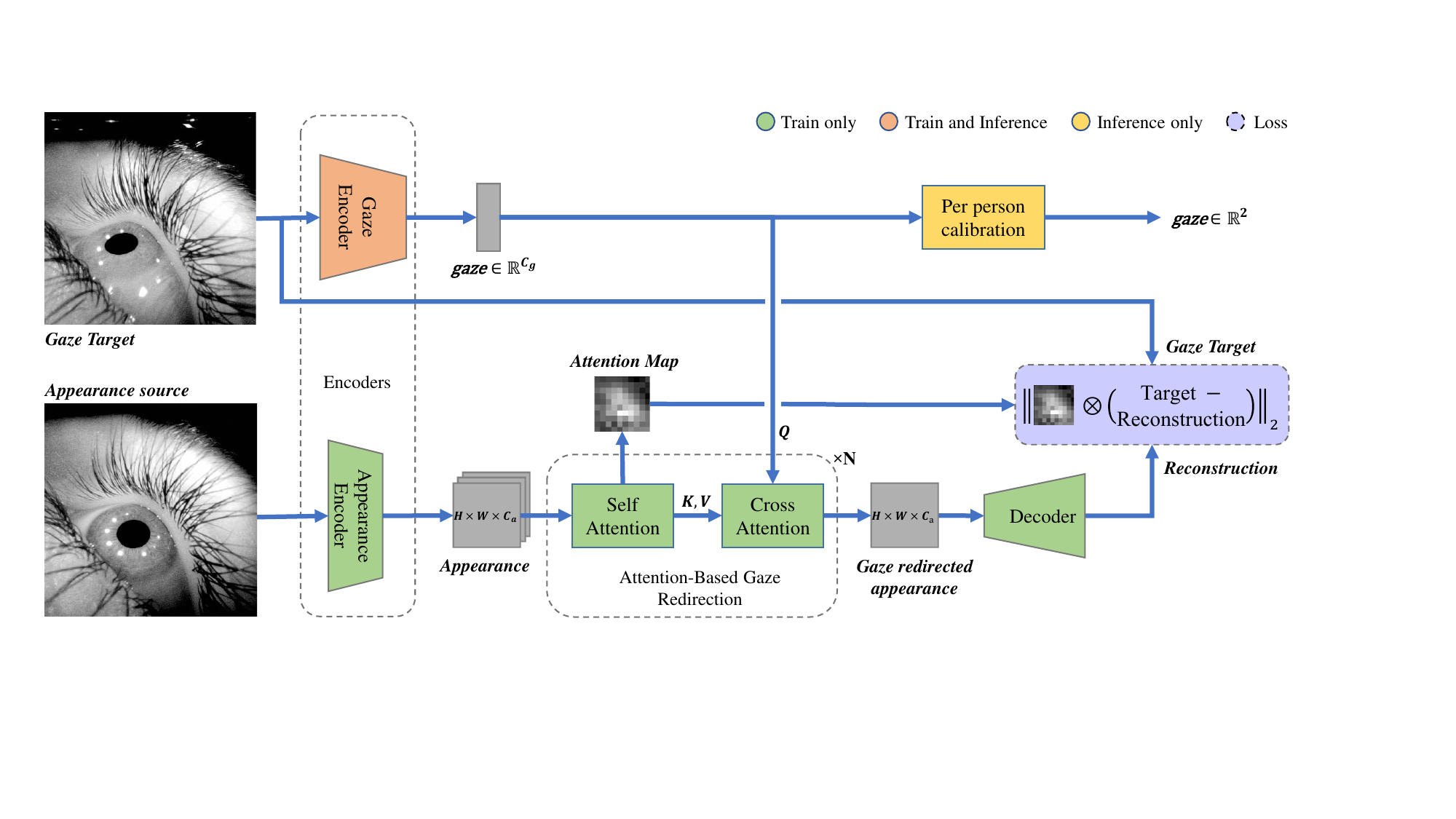}
\centering
\caption{
\textbf{GazeShift architecture.} The model is trained to reconstruct the appearance source image to match the gaze target, conditioned on the target's gaze embedding. A lightweight gaze encoder extracts the embedding, while a separate appearance encoder preserves spatial structure. Appearance tokens undergo self-attention and cross-attention (conditioned on the gaze embedding) before being decoded. Attention maps also guide a gaze-aware loss focused on gaze-relevant regions. At inference, only the gaze encoder and a lightweight calibration module are used to predict gaze.
}
\label{fig:architecture}
\end{figure*}

\section{Introduction}

Gaze estimation is a key component in human–computer interaction \cite{majaranta2014eye}, extended reality (XR) \cite{pfeuffer2017gaze}, and assistive technologies \cite{8856446}. It enables hands-free interaction \cite{sidenmark2021radi}, attention-aware interfaces \cite{mania2021gaze}, and personalized user experiences. In XR, gaze supports foveated rendering \cite{patney2016foveated}, intuitive input, and adaptive content delivery \cite{duchowski2017eye,pfeuffer2017gaze}, with further use in driver monitoring \cite{lian2024driving} and attention analysis \cite{munoz2022discovering}. 

Training accurate gaze estimators requires large datasets with precise labels, yet gaze cannot be directly inferred from eye images. Labels derived from predefined targets are often unreliable due to fixation uncertainty and involuntary saccades, making annotation time-consuming and error-prone \cite{zhang2017mpiigaze}.

Unsupervised learning, successfully applied to classification \cite{NEURIPS2020_f3ada80d,caron2021emerging,he2020momentum}, has been recently explored for gaze estimation \cite{sun2021cross,yu2020unsupervised} to eliminate label dependence. However, existing methods are designed for remote RGB cameras that capture full faces and have not been adapted to the distinct near-eye infrared modality of VR.

To address this gap, we introduce VRGaze, the first large-scale near-eye off-axis gaze dataset, comprising 2.1M labeled images from 68 subjects. Off-axis cameras are standard in modern VR headsets, yet no public dataset previously captured this geometry at comparable scale.

We further propose GazeShift, an unsupervised framework tailored to near-eye imaging but generalizable to remote settings. Unlike prior redirection-based methods relying on geometry \cite{wei2025gazegaussian} or complex warping fields \cite{yu2020unsupervised}, GazeShift learns gaze representations through attention-driven transformations between pairs of the same eye. In VR-mounted cameras, most appearance variation arises from gaze changes; GazeShift models this relation via cross-attention between separately encoded latent representations, enabling clean gaze–appearance disentanglement without geometric priors \cite{sun2021cross,yu2020unsupervised}.

Head-mounted cameras inherently suppress external variation, making gaze the dominant source of temporal change. GazeShift leverages this property with an attention-based reconstruction module producing spatial maps that highlight gaze-relevant regions. To suppress peripheral reconstruction noise, we introduce a gaze-aware loss that focuses learning on regions contributing most to gaze estimation. This creates a feedback loop where refined attention improves performance, which in turn sharpens attention localization.

Evaluated on VRGaze, GazeShift achieves a mean error of $1.84^{\circ}$, nearing supervised accuracy, and generalizes to remote settings, reaching $7.15^{\circ}$ on MPIIGaze with 10× fewer parameters, suitable for deployment on resource-limited headsets.

Our main contributions are:
    
\begin{itemize}

\item The first large-scale off-axis VR gaze dataset (\textbf{VRGaze}) comprising 2.1M labeled images from 68 subjects.

\item An unsupervised framework that learns gaze redirection through cross-image attention and introduces a gaze-aware loss derived from internal attention, focusing learning on gaze-relevant regions without external supervision.

\item Demonstrated state-of-the-art accuracy and cross-modality generalization across VR and remote camera benchmarks. Underscoring its efficiency for edge deployment, the system achieves real-time 5 ms inference on a VR headset with 10x fewer parameters and 35x fewer FLOPs than the baseline.

% \item State-of-the-art performance on both VR and remote camera benchmarks, demonstrating strong cross-modality generalization with $\times10$ fewer parameters and $\times35$ fewer FLOPs than the remote baseline and real-time 5 [ms] inference on a VR headset, underscoring its efficiency for edge deployment.
\end{itemize}

\section{Related Work}

% \begin{table}[t]
% \centering
% \small
% \begin{tabular}{lcccc}
% \toprule
% Method & Eye-only & Shared Enc. & Explicit Warping & Attn-weighted Loss \\
% \midrule
% Yu \& Odobez (2020) & \checkmark & \checkmark & \checkmark & \xmark \\
% Cross-Encoder (2021) & \checkmark & \checkmark & \xmark & \xmark \\
% Equivariance (multi-view) & \xmark & varies & \xmark & \xmark \\
% \textbf{GazeShift (ours)} & \checkmark & \textbf{\xmark} & \textbf{\xmark} & \textbf{\checkmark} \\
% \bottomrule
% \end{tabular}
% \caption{Conceptual differences. GazeShift replaces shared encoders and geometric warping with a minimal attention-only redirection and an attention-weighted loss.}
% \label{tab:novelty-contrast}
% \end{table}

%Cropped eyes vs. full face 
\subsection{Supervised Gaze Estimation}
\textbf{Near-eye cameras.}
% Supervised gaze estimation methods in near-eye settings can be broadly divided into three categories:
Supervised gaze estimation methods for near-eye cameras can be broadly categorized into \textit{appearance-based}, \textit{feature-based}, and \textit{geometry-based} approaches. Appearance-based methods~\cite{kim2019nvgaze} employ end-to-end deep models to directly regress gaze direction from raw eye images. Feature-based methods~\cite{oh2022improved} first extract cues such as the pupil center and corneal reflections (glints) before estimating gaze in a subsequent regression stage. Geometry-based methods~\cite{kaur2022rethinking, zhou2017two} instead rely on explicit eye models and calibrated setups; while highly accurate, they depend on specialized hardware and detailed knowledge of the 3D camera–illumination configuration, which is often unavailable in commercial VR/XR headsets.
% \begin{enumerate}
%     \item Appearance-based approaches~\cite{kim2019nvgaze}, which employ end-to-end deep learning models to directly regress gaze direction from raw eye images.
%     \item Feature-based approaches \cite{oh2022improved}, which first extract eye features such as the pupil center and corneal reflections (glints), and then estimate gaze in a subsequent regression stage.
%     \item Geometry-based approaches \cite{kaur2022rethinking, zhou2017two}, which use explicit eye models and calibrated setups to compute gaze direction. Although these methods can achieve high accuracy, they depend on specialized calibrated hardware, making them impractical for VR/XR applications where the 3D configuration of cameras and illumination sources is not disclosed by headset manufacturers.
% \end{enumerate}

% Gaze estimation 
% methods in near-eye settings are typically categorized as geometry-based or appearance-based. Geometry-based approaches~\cite{kaur2022rethinking, zhou2017two} rely on explicit eye models and calibrated setups to compute gaze direction. While accurate, they require specialized calibrated hardware, which limits their practicality in VR/XR applications. In contrast, appearance-based methods~\cite{kim2019nvgaze} use deep learning to directly infer gaze from images, making them more suitable for consumer-grade headsets. 

\parvspace
\paragraph{Remote cameras.} Early gaze estimation methods focused on predicting gaze direction from eye-region images alone~\cite{cheng2020gaze, park2018deep}. Zhang et al.~\cite{zhang2017s} showed that full-face images could outperform eye-only approaches, leading subsequent research to favor face-based models~\cite{bao2021adaptive}. As a result, most neural network–based methods in remote settings now rely on face images as their primary input – limiting their applicability to near-eye camera scenarios, where only partial facial context is available.

\subsection{Unsupervised Gaze Estimation}

Unsupervised learning reduces annotation cost by leveraging proxy tasks. Yu \& Odobez \cite{yu2020unsupervised} first introduced gaze representation learning through gaze redirection, but their model depends on complex warping fields and geometric priors. In contrast, GazeShift achieves similar redirection behavior using standard attention modules, with no task-specific components.
Sun et al.~\cite{sun2021cross} introduced Cross-Encoder, which jointly encodes gaze and appearance into a shared latent space. While effective, this design exposes the decoder to target appearance information, leading to partial leakage and limited disentanglement. In contrast, GazeShift employs two dedicated encoders, one for gaze and one for appearance, and connects them through a cross-attention mechanism. The gaze vector modulates the source appearance features via attention, effectively acting as a buffer layer that isolates the gaze encoder from the decoder and prevents appearance information from contaminating the gaze representation.

% Sun et al. \cite{sun2021cross} proposed Cross-Encoder, which jointly encodes gaze and appearance in a shared latent space, exposing the decoder to target appearance and limiting disentanglement. GazeShift instead uses separate encoders and clean cross-attention conditioning, enabling a compact gaze encoder for real-time inference.

Other recent unsupervised frameworks \cite{jindal2023contrastive, gideon2022unsupervised, bsautoencoders} employ contrastive or equivariant objectives on full-face inputs, making them less applicable to headset-mounted cameras. GazeShift addresses this gap by learning directly from single-eye infrared imagery.

\section{VRGaze Dataset}\label{sec:vrgaze}

% Despite the growing popularity of VR and the central role of gaze estimation within it, only a few publicly available datasets support gaze estimation in VR settings. OpenEDS2020~\cite{palmero2020openeds2020} includes 550{,}400 gaze-labeled images from 80 participants; however, the data consists of on-axis eye images, which do not reflect the off-axis camera geometry typical of commercial VR headsets.

\begin{figure*}[t]
\includegraphics[clip, trim=7cm 0cm 0cm 0cm,width=0.8\linewidth]{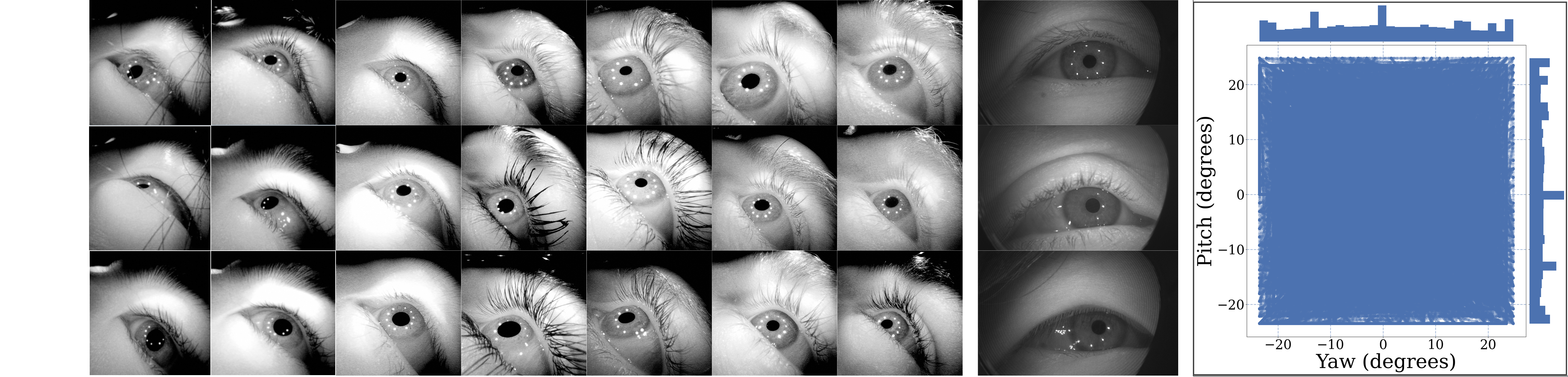}
\centering
\caption{Sample images from our off-axis VRGaze dataset (left), example images from on-axis OpenEDS2020 dataset (center), 2D gaze angle distribution of VRGaze data (right).}
\label{fig:datasample}
\end{figure*}

Despite the growing popularity of VR and the central role of gaze estimation within it, only a limited number of publicly available datasets address this task in VR settings. OpenEDS2020~\cite{palmero2020openeds2020} provides 550{,}400 gaze-labeled images from 80 participants; however, all images are captured with on-axis cameras, which do not reflect the off-axis geometry typical of commercial VR headsets (Apple Vision Pro, Meta Quest Pro, HTC VIVE Pro Eye etc.) \cite{apple2020eyetracking, meta_questpro_eye_tracking_whitepaper, viveproeye_specs}, where cameras are mounted at oblique angles to reduce visual obstruction. This geometry introduces strong perspective distortions absent from on-axis datasets such as OpenEDS, making off-axis data essential for realistic VR gaze estimation (Figure \ref{fig:datasample}). In our experiments (section \ref{par:on-axis_val}), we qualitatively show that on-axis data fail to transfer well to off-axis scenarios, underscoring the importance of our off-axis VRGaze dataset.

% Despite the growing popularity of VR and the central role of gaze estimation within it, only a limited number of publicly available datasets support this task in VR settings. OpenEDS2020~\cite{palmero2020openeds2020} contains 550{,}400 gaze-labeled images from 80 participants; however, all images are captured with on-axis cameras, which do not reflect the off-axis geometry typical of commercial VR headsets. We validate this limitation quantitatively in Section \ref{par:on-axis_val}.

% by training a model on the on-axis OpenEDS2020 dataset and demonstrating that it fails to transfer effectively to the off-axis VRGaze setting

NVGaze~\cite{kim2019nvgaze} offers a larger corpus of 2.5 million frames, yet the majority are also on-axis. Only a small subset of approximately 260K frames from 14 subjects contains off-axis views, and these lack diversity in gaze directions. TEyeD \cite{TEyeD} is a large dataset of over 20 million eye images, but most of it was not collected in a VR environment. Additionally, it was annotated using computational methods and may not provide accurate ground truth. 
These differences are summarized in Table ~\ref{tab:vr_datasets}.
%A summary of the relevant datasets is presented in Table ~\ref{tab:vr_datasets}.

% \begin{table}[t!]
% \caption{Overview of VR eye-tracking datasets.}
% \label{tab:vr_datasets}
% \centering
% \scriptsize % (use \footnotesize if you prefer larger text)
% \begingroup
% \setlength{\tabcolsep}{3pt} % tighten inter-column spacing locally
% \resizebox{0.9\columnwidth}{!}{%
% \begin{tabularx}{\columnwidth}{@{}Y c c r@{\,/\,}r r@{\,/\,}r@{}}
%     \toprule
%     Dataset & Year & Gaze Ann. &
%     \multicolumn{2}{c}{Participants} &
%     \multicolumn{2}{c}{Images} \\
%     \cmidrule(lr){4-5} \cmidrule(lr){6-7}
%      & & (PoR) & VR off-axis & Total & VR off-axis & Total \\
%     \midrule
%     \makecell[l]{NVGaze\\\cite{kim2019nvgaze}} & 2018 & Yes & 14 & 49 & 260K & 2.5M \\
%     \makecell[l]{OpenEDS2020\\\cite{palmero2020openeds2020}} & 2020 & Yes & 0 & 80 & 0 & 550K \\
%     \makecell[l]{TEyeD\\\cite{TEyeD}} & 2021 & No & 14 & 132 & 260K & 20M \\
%     \makecell[l]{VRGaze (ours)} & 2025 & Yes & 68 & 68 & 2.1M & 2.1M \\
%     \bottomrule
% \end{tabularx}
% }
% \endgroup
% \end{table}

\begin{table}[t!]
\caption{Overview of VR eye-tracking datasets.}
\label{tab:vr_datasets}
\centering
\scriptsize % (use \footnotesize if you prefer larger text)
\begingroup
\setlength{\tabcolsep}{3pt} % tighten inter-column spacing locally
\resizebox{1.0\columnwidth}{!}{%
\begin{tabularx}{\columnwidth}{@{}Y c c r@{\,/\,}r r@{\,/\,}r@{}}
    \toprule
    Dataset & Year & Gaze Ann. &
    \multicolumn{2}{c}{Participants} &
    \multicolumn{2}{c}{Images} \\
    \cmidrule(lr){4-5} \cmidrule(lr){6-7}
     & & (PoR) & VR off-axis & Total & VR off-axis & Total \\
    \midrule
    NVGaze \cite{kim2019nvgaze} & 2018 & Yes & 14 & 49 & 260K & 2.5M \\
    OpenEDS2020 \cite{palmero2020openeds2020} & 2020 & Yes & 0 & 80 & 0 & 550K \\
    TEyeD \cite{TEyeD} & 2021 & No & 14 & 132 & 260K & 20M \\
    VRGaze (ours) & 2025 & Yes & 68 & 68 & 2.1M & 2.1M \\
    \bottomrule
\end{tabularx}
}
\endgroup
\end{table}

\begin{table}[t!]
\caption{VRGaze dataset statistics.}
\label{tab:data_span}
\centering
\scriptsize % or \footnotesize if you prefer
\begingroup
\setlength{\tabcolsep}{5pt} % reduce inter-column padding locally
\resizebox{1.0\columnwidth}{!}{%
\begin{tabularx}{\columnwidth}{@{}l c c c@{\:\:/\:\:}c@{\:\:/\:\:}c@{\:\:/\:\:}c c@{}}
    \toprule
    & Gender & Ethnicity
    & \multicolumn{4}{c}{Age (years)}
    & \multirow{2}{*}{Train / Val} 
    \\
    & (F / M) & (Asian / Cauc)     
    & 18–30 & 31–40 & 41–50 & 51+
    \\
    \midrule
    Count 
      & 14 / 54   % Gender F/M
      & 29 / 39   % Ethnicity 
      & 22 & 30 & 13 & 3   % Age bins
      & \, 61 / 7     \\      % Train/Val
    \bottomrule
\end{tabularx}
}
\endgroup
\end{table}

% \begin{figure}[t!]
% \includegraphics[clip, trim=0cm 0cm 0cm 0cm,width=1.0\linewidth]{figures/pupils_dilation_illustration1.png}
% \centering
% \caption{Pupil dilation illustration  }
% \label{fig:pupil_dilation}
% \end{figure}

% \begin{wrapfigure}{r}{0.39\linewidth}
% \vspace{-12pt}
% \includegraphics[clip, trim=0cm 0cm 0cm 0cm,width=1.0\linewidth]{figures/data_statistics4.png}
% \centering
% \caption{2D gaze angle distribution for all VRGaze data}
% \label{fig:datastatistics}
% \vspace{-12pt}
% \end{wrapfigure}

To address the scarcity of large-scale off-axis data, we introduce VRGaze, a new dataset comprising 2.1 million synchronized left and right eye images captured from 68 participants using a custom modern VR headset equipped with off-axis near-eye infrared cameras. Data were recorded at 30 fps with an image resolution of $400\times400$ pixels (sample images are shown in Figure.~\ref{fig:datasample}). The participant pool is diverse in terms of ethnicity, age, and gender (see Table~\ref{tab:data_span}), and all individuals provided informed consent for public release of their eye imagery.

Data were collected as participants followed a moving target on a VR display, alternating between pursuit and fixation segments. The background brightness was varied to elicit a broad range of pupil dilations.

% Data collection involved participants following a moving target displayed on a VR screen. Two target motion patterns were employed: (1) sequences combining smooth pursuit segments followed by fixation, and (2) fixation only intervals where the gaze target remained stationary for three seconds. 
% Additionally, the background brightness was varied from dark to bright to elicit a wide range of pupil dilations. 

Recordings were conducted over multiple sessions. Each participant began with a deterministic target trajectory designed to simulate calibration, followed by sessions featuring randomized target movements to emulate naturalistic viewing behavior. On average, seven sessions were recorded per participant. Seven participants were held out for validation, while the remaining were used for training. 
The distribution of ground-truth gaze angles -- both horizontal (yaw) and vertical (pitch) -- is visualized in Figure ~\ref{fig:datasample}. Gaze labels correspond to the 2D Point of Regard (PoR) of the moving target on the VR display. These coordinates are computed from the known geometry of the headset. For validation, we consider only fixation intervals, where gaze remains stable and PoR annotations are the most reliable.

\section{Method}

We introduce GazeShift, a simple yet general attention-based framework for unsupervised gaze representation learning. Unlike prior gaze-redirection or disentanglement models that depend on geometric priors, multi-view consistency, or complex regularization schemes~\cite{yu2020unsupervised, wei2025gazegaussian, kurzhals2020view}, GazeShift relies solely on standard attention operations to model transformations between eye images captured at different gaze directions. Its core novelty lies in a generic attention architecture free of ad hoc gaze-specific modules, and in a gaze-aware reconstruction loss that leverages the model’s own self-attention maps to automatically emphasize gaze-relevant regions. Together, these components enable effective learning of gaze-specific embeddings without labels, auxiliary detectors, or handcrafted constraints, making GazeShift both conceptually simple and broadly applicable across VR and remote-camera domains. The overall architecture is illustrated in Figure~\ref{fig:architecture}.

% We introduce GazeShift, a simple yet generic attention-based framework for unsupervised gaze representation learning. Unlike prior gaze-redirection or disentanglement models that rely on geometric priors, multi-view consistency, or complex regularization \cite{yu2020unsupervised, wei2025gazegaussian, kurzhals2020view}, GazeShift uses only standard attention operations to model the transformation between eye images captured at different gaze directions. Its core novelty lies in a lightweight, architecture-agnostic formulation and in a gaze-aware reconstruction loss that automatically emphasizes gaze-relevant regions based on attention maps. Together, these components enable effective learning of gaze-specific embeddings without labeled data, auxiliary detectors or task-specific constraints, making GazeShift both conceptually simple and broadly applicable across VR and remote-camera domains. Our architecture is depicted in Figure  \ref{fig:architecture}.

Assuming that, for a given subject and eye, gaze changes are the dominant source of appearance variation across frames, we train a model to transform a source frame into a target frame, conditioned on an embedding extracted from the target. The key idea is that this embedding must encode the information required to redirect the source gaze toward the target gaze. If gaze indeed explains most inter-frame differences, the learned embedding will naturally become rich in gaze-specific cues. To ensure that source–target variation is primarily gaze-related, we form training batches from pairs of frames sampled from the same eye of the same individual at different times.

\subsection{Separate Gaze and Appearance Encoders}
To perform this generative pretext task, we extract appearance information from the source frame and gaze information from the target frame using two separate encoders.

Let \( \mathbf{x}_s \) and \( \mathbf{x}_t \) denote the source and target frames, respectively.  
We extract a source appearance feature map \( \mathbf{A}_s \in \mathbb{R}^{H \times W \times C_a} \) from \( \mathbf{x}_s \) using the appearance encoder \( f_{\text{app}}(\cdot) \), and a target gaze embedding \( \mathbf{g}_t \in \mathbb{R}^{C_g} \) from \( \mathbf{x}_t \) using the gaze encoder \( f_{\text{gaze}}(\cdot) \):

\begin{equation}
\mathbf{A}_s = f_{\text{app}}(\mathbf{x}_s), \quad \mathbf{g}_t = f_{\text{gaze}}(\mathbf{x}_t).
\end{equation}

In contrast to prior work such as Cross-Encoder \cite{sun2021cross}, which uses a single shared encoder for both appearance and gaze, we explicitly separate the two. 
We observe that gaze is an abstract, non-spatial attribute (typically described by 2 or 3 real-valued angles per frame), requiring a deep gaze encoder to match its abstraction level.
In contrast, appearance is a spatial, concrete attribute, closely tied to the local image structure. 
Thus, the appearance encoder is designed to be shallower and preserves the 2D structure of the source frame.
The architectural separation between gaze and appearance not only reflects their fundamentally different nature -- but also facilitates effective disentanglement between them.  
This asymmetry further enables a lightweight gaze encoder and a heavier appearance encoder, with no runtime penalty, as only the gaze encoder is used during inference.

While our framework is architecture-agnostic, we target lightweight, real-time-compatible models to meet the constraints of VR and edge devices.  
In particular, we implement the gaze encoder using inverted bottleneck blocks from MobileNetV2~\cite{sandler2018mobilenetv2}. 
Detailed architectural descriptions of the encoders, attention modules, and decoder are provided in the supplementary material.

\subsection{Gaze-Conditioned Global Modulation}

The appearance encoder output, $\mathbf{A}_s$, is first refined via multi-head self-attention to capture spatial interactions, producing updated features $\mathbf{A}_s' \in \mathbb{R}^{H \times W \times C_a}$:
\begin{equation}
\mathbf{A}_s' = \text{SelfAttn}(\mathbf{A}_s)
\label{eq:self_attn}
\end{equation}

To condition these features on the target gaze without disrupting their spatial structure, we employ a global modulation strategy. The target gaze embedding $\mathbf{g}_t \in \mathbb{R}^{C_g}$ is linearly projected to dimension $C_a$ and used as a single global query. Cross-attention is then applied with this gaze query and the spatial appearance features $\mathbf{A}_s'$ as keys and values, yielding a gaze-conditioned global context vector $\mathbf{c} \in \mathbb{R}^{C_a}$:
\begin{equation}
\mathbf{c} = \text{CrossAttn}(\mathbf{q}_g, \mathbf{A}_s', \mathbf{A}_s')
\label{eq:cross_attn}
\end{equation}
where $\mathbf{q}_g \in \mathbb{R}^{C_a}$ denotes the projected gaze query. This global context is then broadcast across the spatial dimensions and added to the appearance features via a residual connection, yielding the fused representation $\mathbf{F}$:
\begin{equation}
\mathbf{F} = \mathbf{A}_s' + \mathbf{C},
\label{eq:fusion}
\end{equation}
where $\mathbf{C} \in \mathbb{R}^{H \times W \times C_a}$ is obtained by spatially broadcasting $\mathbf{c}$.
This residual addition acts as a feature-wise global modulation: it steers the latent representation toward the target gaze direction while preserving the structural integrity and spatial resolution of the original appearance map.

\subsection{Gaze-Focused Reconstruction Loss}

In a standard gaze-redirection task, the model reconstructs the target image from the source using a per-pixel mean squared error loss.
This uniform weighting treats all pixels equally, forcing the model to also reconstruct regions unrelated to gaze, such as background or boundary areas.
As a result, the learned gaze embedding may encode unnecessary appearance details that degrade its gaze specificity.
To mitigate this issue, previous works have relied on externally defined gaze masks or handcrafted geometric priors \cite{yu2020unsupervised, jiang2024learning}.
In contrast, we exploit the model’s own self-attention maps to derive adaptive spatial weights.
Since in our setting most appearance variation between source and target stems from gaze changes, these attention maps naturally highlight gaze-relevant regions.
We therefore use them as soft masks, emphasizing gaze-related pixels while down-weighting non-gaze regions, encouraging the model to focus reconstruction on the eye region that truly conveys gaze direction.

% Leveraging the cross-attention module that modulates source appearance with target gaze, we introduce a gaze-focused reconstruction loss that exploits these attention maps to guide learning toward the most gaze-informative regions. Unlike prior gaze-redirection methods that depend on externally defined gaze masks or handcrafted priors (\cite{yu2020unsupervised, jiang2024learning}), we introduce a gaze-aware reconstruction loss that automatically focuses learning on gaze-relevant regions. The loss derives its spatial weighting directly from the model’s own attention maps, eliminating the need for additional supervision or geometric constraints. This design creates a self-reinforcing loop: as the attention maps become more localized around gaze-sensitive regions (e.g., the iris and pupil boundary), the loss further sharpens supervision in those areas, which in turn improves the accuracy of the learned attention. This mechanism provides an elegant, unsupervised way to emphasize gaze-critical features while ignoring peripheral variations such as eyelids or illumination changes.

Since the attention weight map \( \mathbf{w} \) is computed at a lower resolution due to encoder downsampling, we first upsample \( \mathbf{w} \in \mathbb{R}^{H \times W} \) to match the spatial resolution of the target \( \mathbf{x}_t \) and reconstructed frame \( \hat{\mathbf{x}}_t \). The \textit{gaze-focused reconstruction loss} with a sharpening parameter \( \gamma > 0 \) is defined as:

\begin{equation}
\mathcal{L}_{\text{focus}} = \frac{1}{\sum_{i} w_i^\gamma} \sum_{i} w_i^\gamma \left( x_{t,i} - \hat{x}_{t,i} \right)^2
\end{equation}

where \( i \) indexes pixel locations, \( w_i \) denotes the upsampled attention weight at pixel \( i \), and \( \gamma \) controls the sharpness of the focus.

Setting $\gamma=1$ recovers the standard attention-weighted loss, while $\gamma>1$ sharpens the attention map, increasing focus on gaze-relevant regions, and $\gamma<1$ softens it, distributing supervision more broadly. This self-supervised weighting scheme inherently couples attention maps with gaze representation fidelity, eliminating the need for additional regularization. Formally, for $L_{\text{focus}}=\sum_i \tilde{w}_i(\hat{x}_i-x_i)^2$ with $\tilde{w}_i=\frac{w_i^\gamma}{\sum_j w_j^\gamma}$ and $\gamma>0$, the per-pixel gradient magnitude scales as $\left\lVert \frac{\partial L_{\text{focus}}}{\partial \hat{x}_i} \right\rVert \propto \tilde{w}_i = \frac{w_i^\gamma}{\sum_j w_j^\gamma}$. Increasing $\gamma$ thus amplifies gradients in high-attention (gaze-relevant) regions and suppresses them elsewhere, providing a self-guided mechanism that focuses learning on gaze cues while reducing overfitting to peripheral variations.

\subsection{Gaze Calibration}
% After unsupervised pretraining, the model outputs latent embeddings encoding gaze direction.
After unsupervised pretraining, the gaze encoder outputs latent embeddings encoding gaze direction.
To derive the final gaze estimation, calibration is performed differently in near-eye and remote-camera configurations.
In the VR setting, where high precision is required, we apply a lightweight few-shot calibration akin to adaptation steps in self-supervised learning.
A small set of labeled fixation points is used to fit a linear regressor mapping embeddings to 2D gaze angles.
Because the \textit{kappa angle}, the offset between optical and visual axes, varies across individuals, per-person calibration is performed.
Additionally, session-wise calibration before each recording compensates for headset slippage or repositioning, ensuring consistent gaze alignment.

For remote-camera datasets, no user-specific fitting is applied.
Instead, a small MLP regressor is trained on a shared pool of 100–200 labeled samples aggregated across all subjects, following \cite{sun2021cross, yu2020unsupervised}.

\section{Experiments}
%We evaluate GazeShift on both remote-camera and near-eye camera datasets. %Our goals are to assess (1) unsupervised representation quality, (2) cross-domain generalization, and (3) efficiency for real-time inference.

To evaluate the effectiveness of GazeShift, we conduct experiments in two settings: a \textit{near-eye VR} setup using our newly collected VRGaze off-axis dataset and the smaller OpenEDS2020 dataset~\cite{palmero2020openeds2020} (550K on-axis eye images), and a \textit{remote-camera} setup using the Columbia~\cite{smith2013gaze} and MPIIGaze~\cite{zhang2017mpiigaze} datasets.

% To assess the effectiveness of GazeShift, we perform experiments across two settings:
% \begin{enumerate}
%     \item \textbf{Near-eye VR:} using our newly collected VRGaze off-axis camera dataset and the smaller OpenEDS2020 dataset \cite{palmero2020openeds2020}, which comprises 550K on-axis eye images.
%     \item \textbf{Remote camera:} using the Columbia \cite{smith2013gaze} and MPIIGaze \cite{zhang2017mpiigaze} datasets.
% \end{enumerate}
For the off-axis VR setting, we compare GazeShift against Cross-Encoder~\cite{sun2021cross}, the current leading unsupervised gaze-estimation baseline for eye-only imagery. While an earlier method by Yu and Odobez~\cite{yu2020unsupervised} also explores unsupervised gaze learning for eye images, its source code is not publicly available and its reported results are notably inferior to Cross-Encoder. Consequently, Cross-Encoder remains the only reproducible and competitive unsupervised method applicable to eye-only input, making it the most relevant baseline for this setting. (Other unsupervised approaches either require full-face or multi-view input, depend on geometric priors, or lack publicly available implementations, rendering them unsuitable for direct comparison in the off-axis VR setting.)

All training details, including hyperparameters and implementation specifics, are provided in the appendix.

\subsection{VR Experiments}

% \textbf{Supervised model.}
% To compare GazeShift with supervised methods, we trained two models following widely adopted approaches in gaze estimation: 
% \begin{enumerate}
%     \item Appearance-based model – implemented as a binocular Siamese network that fuses features from the left and right eye images to predict gaze direction.
%     \item Feature-based model – trained to regress pupil center coordinates and glint positions. The predicted features are then mapped to gaze labels using a linear fitting procedure.
% \end{enumerate}

\textbf{Supervised model.}
To compare GazeShift with supervised methods, we train two supervised baselines following standard gaze-estimation paradigms: an \textit{appearance-based} binocular Siamese network that fuses left- and right-eye features to predict gaze direction, and a \textit{feature-based} model that regresses pupil centers and glint positions, mapping the predicted features to gaze labels through linear fitting.

% \textbf{Supervised model.}
% To compare GazeShift with supervised methods, we trained two models following widely adopted approaches in gaze estimation: 
% \begin{enumerate}
%     \item Appearance-based model – implemented as a binocular Siamese network that fuses features from the left and right eye images to predict gaze direction.
%     \item Feature-based model – trained to regress pupil center coordinates and glint positions. The predicted features are then mapped to gaze labels using a linear fitting procedure.
% \end{enumerate}

% The complete model architecture and training details are provided in the appendix.

\newcolumntype{Y}{>{\centering\arraybackslash}X}
\begin{table}[t]
\caption{Performance on VRGaze.}
\label{table:VRGaze_results}
\centering
\small
\resizebox{0.9\columnwidth}{!}{%
\begin{tabularx}{\columnwidth}{@{}l l X c@{}}
\toprule
Supervision & Method & \makecell[l]{Calibration \\Type} & \makecell{Avg. \\Error [$^\circ$]} \\ 
\midrule
\multirow{2}{*}{Supervised}  & Appearance Based & \multirow{5}{*}{\makecell[l]{Per-person}} & \textbf{1.54} \\ 
            & Feature Based & & 3.2 \\ 
\cmidrule(r){1-2} \cmidrule(l){4-4}
\multirow{7}{*}{Unsupervised} 
            & VAE            &            & 5.30 \\
            & Cross-Encoder  &            & 2.15 \\
            & GazeShift      &            & \textbf{1.84} \\ 
\cmidrule(l){2-4}
\rule{5pt}{0.33\baselineskip}
            & Cross-Encoder  & \multirow{2}{*}{\makecell[l]{Person-agnostic \\(K=100)}} & 2.83 \\
            % \rule{0pt}{0.33\baselineskip}
            & GazeShift      &                          & \textbf{2.73} \\ 
            % \rule{0pt}{0.33\baselineskip}
\cmidrule(l){2-4}
            & Cross-Encoder  & \multirow{2}{*}{\makecell[l]{Person-agnostic \\(K=200)}} & 2.26 \\
            & GazeShift      &                          & \textbf{2.13} \\ 
\bottomrule
\end{tabularx}
}
\end{table}

\begin{table}[t!]
    \centering
    \tiny
    \caption{Performance on OpenEDS2020.}
    \label{tab:OpenEDS2020}
    \resizebox{0.9\columnwidth}{!}{%
    \begin{tabular}{llc}
        \toprule
        Model & Calibration Type & \makecell{Avg. \\Error [°]} \\
        \midrule
        Cross-Encoder & Per-person & 3.69 \\
        GazeShift     & Per-person & \textbf{3.43} \\
        \midrule
        Cross-Encoder & Person-agnostic & 5.20 \\
        GazeShift     & Person-agnostic & \textbf{4.20} \\
        \bottomrule
    \end{tabular}
    }
\end{table}

%%%%%%%%%%%%%%%%%%%%%%%%%%%%%%%%%%%%
% \begin{table}[t]
% \caption{Performance on VRGaze and OpenEDS2020 (Avg. angular error in degrees).}
% \label{table:combined_results}
% \centering
% \small
% \resizebox{\columnwidth}{!}{%
% \begin{tabular}{llccc}
% \toprule
% Supervision & Method & Calibration 
% & VRGaze [$^\circ$] 
% & OpenEDS2020 [$^\circ$] \\
% \midrule

% \multirow{2}{*}{Supervised}
% & Appearance Based & Per-person & \textbf{1.54} & --- \\
% & Feature Based & Per-person & 3.2 & --- \\
% \cmidrule(lr){1-5}

% \multirow{9}{*}{Unsupervised}
% & VAE & Per-person & 5.30 & --- \\
% & Cross-Encoder & Per-person & 2.15 & 3.69 \\
% & GazeShift & Per-person & \textbf{1.84} & \textbf{3.43} \\
% \cmidrule(lr){2-5}

% & Cross-Encoder & Person-agnostic & 2.83 & 5.20 \\
% &  & (K=100) &  &  \\
% & GazeShift & Person-agnostic & \textbf{2.73} & \textbf{4.20} \\
% &  & (K=100) &  &  \\
% \cmidrule(lr){2-5}

% & Cross-Encoder & Person-agnostic & 2.26 & --- \\
% &  & (K=200) &  &  \\
% & GazeShift & Person-agnostic & \textbf{2.13} & --- \\
% &  & (K=200) &  &  \\

% \bottomrule
% \end{tabular}%
% }
% \end{table}
% %%%%%%%%%%%%%%%%%%%%%%%%%%%%%%%%%%%%

\begin{table}[t]
    \centering
    \caption{GazeShift ablation study on VRGaze.}
    \label{table:ablations}
    \small
    \resizebox{0.9\columnwidth}{!}{%
    \begin{tabularx}{\columnwidth}{c *{3}{X} c}
        \toprule
        \# & \makecell[l]{Separate \\Encoders} & \makecell[l]{Attention-\\Based \\Redirection} & 
             \makecell[l]{Gaze-\\Focused \\Loss} & \makecell{Avg. \\Error [°]} \\
        \midrule
        1 & \texttimes & \texttimes & \texttimes & 2.15 \\
        2 & \checkmark & \texttimes & \texttimes & 2.10 \\
        3 & \checkmark & \checkmark & \texttimes & 2.07 \\
        4 & \checkmark & \checkmark & \checkmark & \textbf{1.84} \\
        \bottomrule
    \end{tabularx}
    }
\end{table}

\parvspace
\paragraph{Calibration protocol.}
Calibration samples are extracted from fixation intervals by selecting the frame whose gaze embedding was closest to the median within each interval. Each selected sample consists of a synchronized left/right image pair and a 2D point of regard label. Calibration and evaluation samples are strictly disjoint.

To evaluate gaze estimation accuracy, we map the unsupervised gaze embeddings to 2D gaze labels using ridge regression under two settings:

(1) Per-person: for each subject in the validation set, $K \in \{17, 30, 40, 50, 60\}$ fixation points are randomly sampled for calibration, and the remaining fixation points are used exclusively for evaluation. Results are averaged over 10 random splits across $K$ values and subjects.

(2) Person-agnostic: a single regressor is trained using $K \in \{100, 200\}$ labeled fixation points pooled across validation subjects, with the remaining samples used for evaluation. Results are averaged over 10 runs.

% \textbf{Calibration protocol.}
% Calibration samples were extracted from fixation intervals by selecting the frame whose embedding was closest to the median within each interval. Each selected sample was paired with a 2D point-of-regard label and synchronized left and right eye images. To evaluate gaze estimation accuracy, we fit the unsupervised gaze embeddings to 2D gaze labels using ridge regression under two calibration settings. In all evaluations, calibration and test samples are strictly disjoint. 
% (1) \textit{Per-person} — a separate regressor is trained per subject. For a given subject, $K$ fixation points are randomly sampled from the validation set for calibration, and the remaining points are used for evaluation. This process is repeated 10 times for $K \in \{17, 30, 40, 50, 60\}$, and we report the average error. 
% (2) \textit{Person-agnostic} — a single regressor is trained on $K \in \{100, 200\}$ labeled samples randomly pooled across subjects, averaged over ten runs, and evaluated on the remaining held-out data.

\parvspace
\paragraph{VRGaze results.}
% The proposed GazeShift framework achieves its best performance on the newly collected VRGaze dataset, which reflects modern off-axis near-eye camera configurations. 
As shown in Table~\ref{table:VRGaze_results}, GazeShift consistently outperforms prior unsupervised baselines across all calibration settings, reaching a mean error of \textbf{1.84$^{\circ}$} under per-person calibration and maintaining competitive accuracy under person-agnostic evaluation. 
These results confirm that the attention-based redirection and gaze-aware loss are particularly effective for handling the geometric distortions and illumination variability characteristic of off-axis headset-mounted sensors. 
% Together with the OpenEDS2020 experiments, they demonstrate that GazeShift generalizes across both off-axis and on-axis camera configurations, underscoring its robustness to diverse VR hardware configurations.

% \textbf{VRGaze results.}
% The proposed GazeShift framework achieves its strongest performance on the newly collected VRGaze dataset, which represents modern off-axis near-eye camera configurations. As shown in Table~\ref{table:VRGaze_results}, GazeShift surpasses prior unsupervised baselines across all calibration settings, achieving a mean error of \textbf{1.84\degree} under per-person calibration and maintaining competitive accuracy under person-agnostic evaluation. These results establish VRGaze as a challenging yet reliable benchmark for off-axis VR gaze estimation and confirm that the attention-based redirection and gaze-aware loss are particularly well-suited for the spatial distortions and illumination patterns characteristic of off-axis headset-mounted sensors. In combination with the OpenEDS2020 results, this demonstrates that GazeShift generalizes effectively across both off-axis and on-axis camera geometries, highlighting its robustness to diverse VR hardware configurations.

\parvspace
\paragraph{OpenEDS2020 results.}
To evaluate GazeShift under on-axis conditions, we trained it on the OpenEDS2020 dataset using Cross-Encoder as the baseline. 
As summarized in Table~\ref{tab:OpenEDS2020}, GazeShift consistently outperforms Cross-Encoder, achieving lower errors in both per-person and person-agnostic ($K{=}100$) calibration protocols. 
These results demonstrate that the proposed attention-based representation and gaze-aware loss transfer effectively across different camera geometries and VR headset designs, highlighting the model’s robustness to geometric and photometric variations in real-world conditions.

% \textbf{OpenEDS2020 results.}
% To verify GazeShift performance on additional scenarios such as on-axis data, we trained it on OpenEDS2020, using Cross-Encoder as the baseline. The results, summarized in Table~4, show that GazeShift consistently outperforms Cross-Encoder, achieving lower errors in both the per-person and person-agnostic (K=100) calibration protocols. This demonstrates that the proposed attention-based representation and gaze-aware loss transfer effectively across camera configurations and VR headsets, highlighting the model’s robustness to geometric and photometric variations inherent in real-world devices.

\parvspace
\paragraph{Cross-Dataset generalization from on-axis to off-axis data.}
\phantomsection
\label{par:on-axis_val}
To assess the importance of off-axis data, we evaluate whether a model trained on existing on-axis datasets can generalize to off-axis conditions. 
As illustrated in Figure~\ref{fig:datasample}, the appearance differences between on-axis and off-axis imagery are substantial, with strong perspective and illumination shifts. 
We train GazeShift on the on-axis OpenEDS2020 dataset and test it on the VRGaze validation set, obtaining a mean error of $5.2^{\circ}$ under the per-person calibration protocol, compared to $1.84^{\circ}$ when trained directly on VRGaze. 
This large gap highlights that models trained on on-axis data fail to capture the geometric distortions characteristic of off-axis near-eye cameras, underscoring the necessity of dedicated datasets such as VRGaze for accurate and robust VR gaze estimation.

% \textbf{Cross-Dataset Generalization from On-Axis to Off-Axis Data.}
% \phantomsection
% \label{par:on-axis_val}
% To further evaluate the value of VRGaze, we examine whether a model trained on existing on-axis datasets can generalize effectively to off-axis data. As shown in Figure \ref{fig:datasample}, the appearance differences between the two settings are substantial. To quantify this gap, we train GazeShift on the on-axis OpenEDS2020 dataset and evaluate it on the VRGaze validation set. Under the per-person calibration protocol, the model reached an error of 5.2°, compared to 1.84° when trained directly on VRGaze. These results underscore the importance of dedicated off-axis datasets such as VRGaze for accurate gaze estimation.

\begin{figure}[t]
    \centering
    \includegraphics[clip,trim=0cm 0cm 0cm 0cm, width=0.65\linewidth]{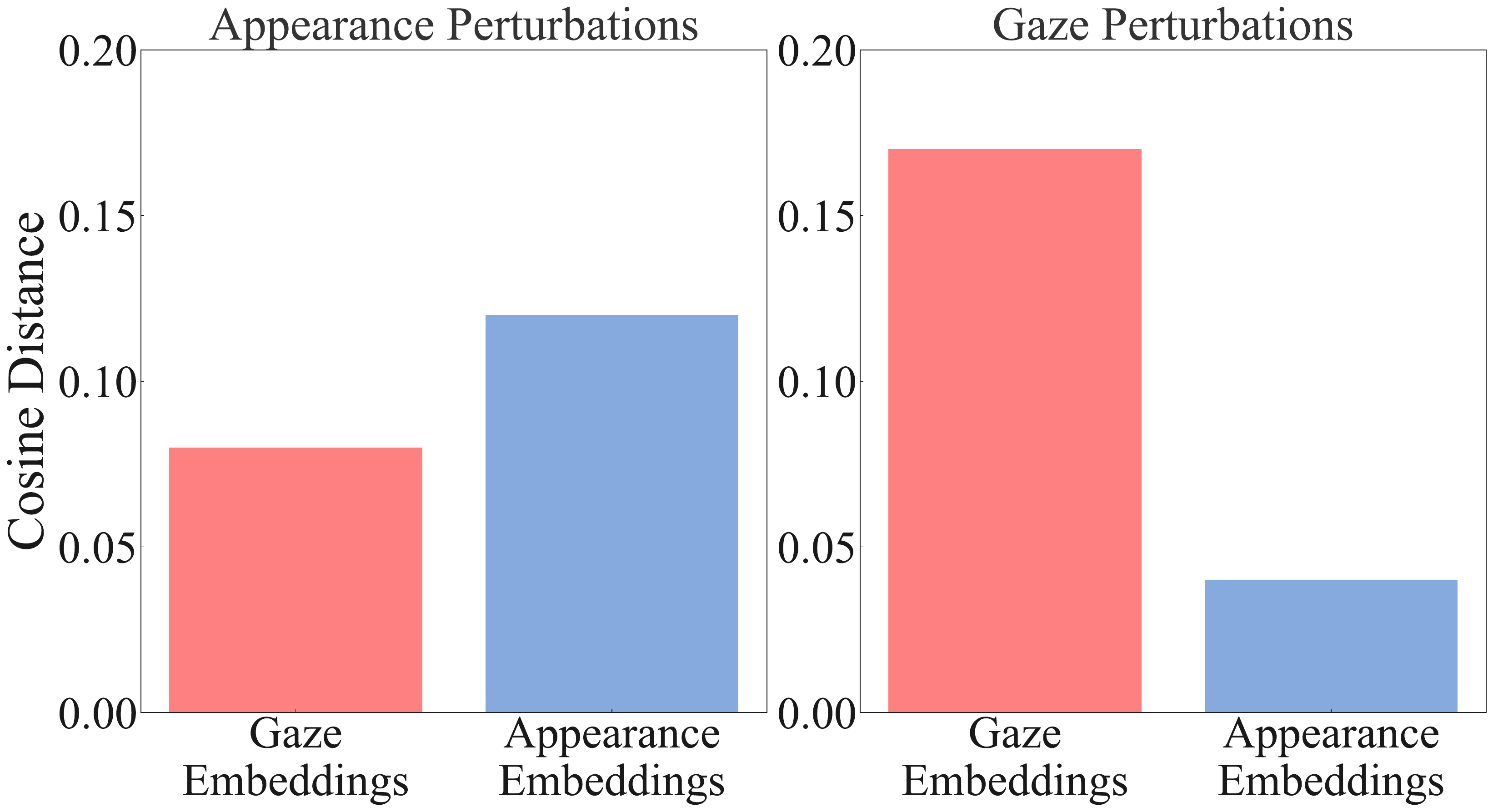}
    \caption{Disentanglement analysis of learned embeddings. Left: appearance perturbations under fixed gaze. Right: gaze perturbations under fixed appearance. Gaze embeddings vary primarily with gaze direction, while appearance embeddings remain comparatively stable, indicating effective disentanglement.}
    \label{fig:distentangle}
\end{figure}

% \begin{figure}[t]
%     \centering
%     \includegraphics[clip,trim=0cm 0cm 0cm 0cm, width=0.7\linewidth]{figures/disentanglement_vector_nogrid_ylim02.pdf}
%     \caption{Disentanglement analysis of learned embeddings. Left: Appearance Perturbations — same gaze, varying appearance. Right: Gaze Perturbations — same appearance, varying gaze.
% Gaze embeddings vary primarily with gaze direction, while appearance embeddings remain stable, demonstrating effective disentanglement.}
%     \label{fig:distentangle}
% \end{figure}

%we designed two complementary analyses (Figure \ref{fig:distentangle}). In the appearance-perturbation experiment, a single eye image was augmented with 100 illumination and contrast variants, keeping gaze fixed. In the gaze-perturbation experiment, 

\parvspace
\paragraph{Validating gaze–appearance disentanglement}
To examine GazeShift's assumption that inter-frame appearance variation primarily reflects gaze shifts, we analyze embedding sensitivity to both gaze and appearance perturbations (Figure \ref{fig:distentangle}). For appearance perturbations, we augment a single eye image with 100 illumination and contrast variations, keeping gaze fixed. For gaze perturbations we sampled 80 images of the same eye with distinct gaze directions but similar lighting. In both experiments, we compute the average cosine distances across all gaze embeddings and all appearance embeddings. Under appearance perturbations, gaze embeddings remained stable (0.08 average cosine distance) while appearance embeddings varied moderately (0.12). Under gaze perturbations, gaze embeddings changed substantially (0.17) whereas appearance embeddings remained nearly constant (0.04). This analysis show that GazeShift disentangles gaze from appearance.

\begin{figure}[t]
    \centering
    \includegraphics[clip, width=0.85\linewidth]{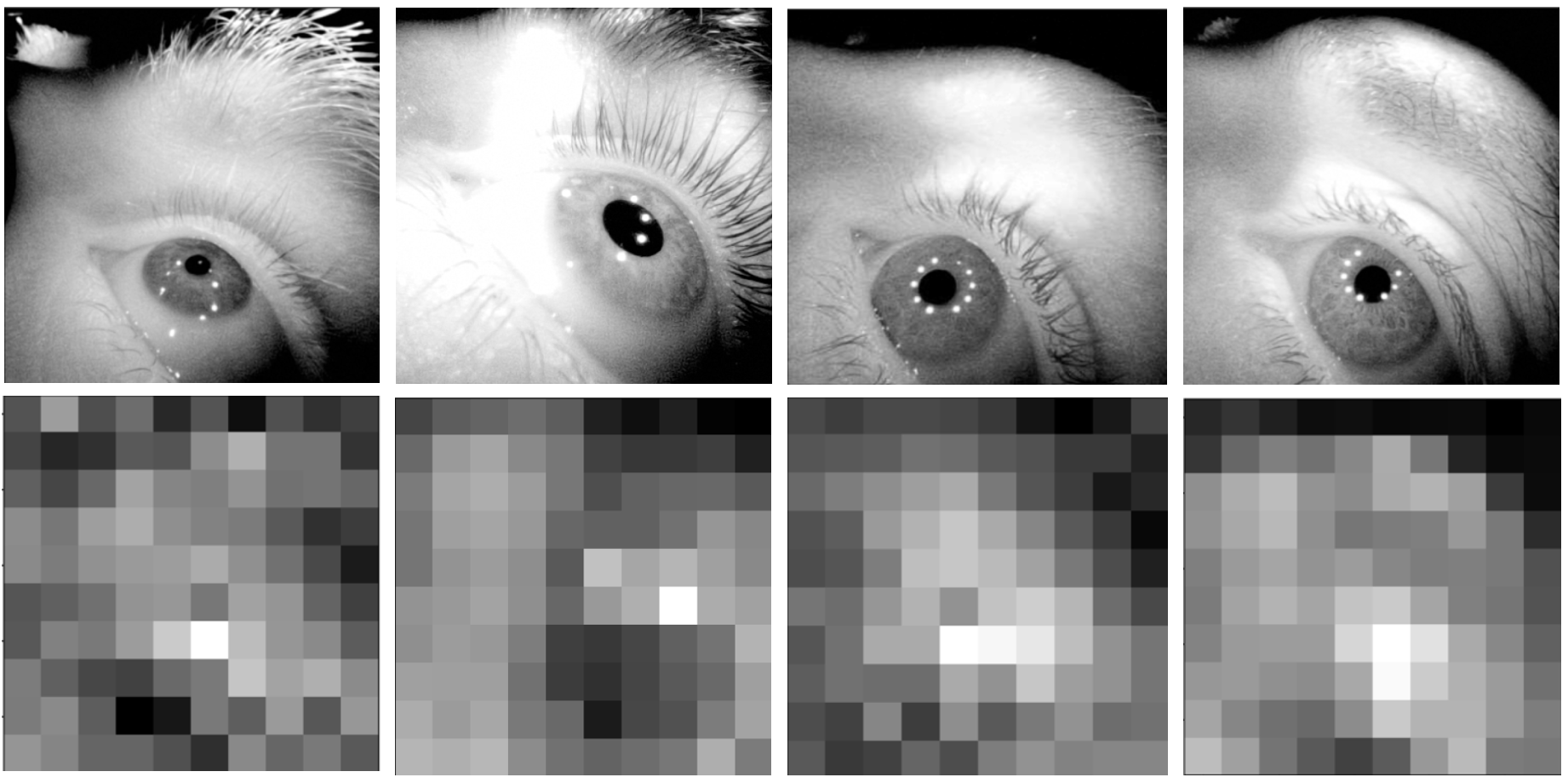}
    \caption{\textbf{Appearance source frames and their associated self-attention maps.} The model learns to focus on regions that exhibit the greatest differences between the source and target images---primarily the gaze-relevant areas around the iris.}
    \label{fig:att_map}
\end{figure}

\parvspace
\paragraph{Attention visualization.}
Figure~\ref{fig:att_map} presents example source appearance images alongside their corresponding self-attention maps. 
The model learns to focus on regions that exhibit the greatest appearance differences between the source and target images, primarily the gaze-relevant areas around the iris.

% Figure~\ref{fig:att_map} presents example source appearance images alongside their corresponding self-attention maps. The model learns to focus on regions that exhibit the greatest appearance differences between the source and target images -- primarily the gaze-relevant areas around the iris.

% \begin{table}[t!]
%     \centering
%     \tiny
%     \caption{Average error under different $\gamma$ settings.}
%     \label{tab:gamma_ablation}
%     \resizebox{0.5\columnwidth}{!}{%
%     \begin{tabular}{lc}
%         \toprule
%         \textbf{$\gamma$} & Avg. Error [°] \\
%         \midrule
%         0.5               & 2.03 \\
%         1.0               & 1.84 \\
%         2.0               & 2.19 \\
%         % Trainable (2.32)  & 2.25 \\
%         4.0               & 2.41 \\
%         \bottomrule
%     \end{tabular}
%     }
% \end{table}

\begin{table}[t]
    \centering
    \tiny
    \caption{Average error under different $\gamma$ settings.}
    \label{tab:gamma_ablation}
    \resizebox{0.8\columnwidth}{!}{%
    \begin{tabular}{lcccc}
        \toprule
        $\gamma$ & 0.5 & 1.0 & 2.0 & 4.0 \\
        \midrule
        Avg. Error [$^\circ$] & 2.03 & \textbf{1.84} & 2.19 & 2.41 \\
        \bottomrule
    \end{tabular}
    }
\end{table}

\parvspace
\paragraph{Model efficiency and on-device runtime.}
Our lightweight gaze encoder contains 342K parameters and requires 55 MFLOPs. To assess deployment feasibility, we implemented GazeShift gaze encoder on our custom VR headset equipped with an Exynos 2200 chipset. Running on the Xclipse 920 mobile GPU, average runtime for both eyes was measured at 5 ms, confirming GazeShift’s suitability for real-time, on-device gaze applications.

\parvspace
\paragraph{Ablations studies.} 
We systematically evaluate each component of the GazeShift framework on the VRGaze dataset, with results summarized in Table~\ref{table:ablations}. The baseline configuration uses a single shared encoder for both gaze and appearance, with concatenated features and a standard $\ell_2$ reconstruction loss, yielding an average error of 2.15$^\circ$. Splitting the encoder into distinct gaze and appearance branches reduces the error to 2.10$^\circ$, confirming the benefit of disentangled representations. Incorporating cross-attention further lowers the error to 2.07$^\circ$, as it enables direct conditioning of appearance features on gaze information. Crucially, this attention mechanism also makes possible the gaze-focused reconstruction loss, which leverages attention maps to weight supervision on gaze-relevant regions, yielding the largest improvement, down to 1.84$^\circ$

% We conduct ablation experiments on the VRGaze dataset to quantify the contribution of each component in GazeShift. Results are summarized in Table 5. The baseline configuration (row 1) uses a single shared encoder for both gaze and appearance, fusing the two representations via concatenation and optimizing a standard $\ell_2$ reconstruction loss, yielding 2.15\degree.

% Introducing separate encoders for gaze and appearance (row 2) reduces the error to 2.10\degree, confirming that explicit feature separation improves representation purity. Incorporating attention-based redirection (row 3) further lowers the error to 2.07\degree, showing that a generic attention mechanism is sufficient to model gaze shifts without specialized geometric priors or warping fields. Finally, adding the gaze-focused loss (row 4) results in a substantial improvement to 1.84\degree, demonstrating that attention-guided supervision significantly enhances the discriminative quality of the learned gaze embeddings.

% Together, these results validate the two central innovations of GazeShift: (1) a lightweight, architecture-agnostic attention framework that captures gaze variation without task-specific modules, and (2) a gaze-aware reconstruction loss that self-adapts supervision to gaze-relevant regions.

\parvspace
\paragraph{Gaze-focused loss sensitivity.} We experimented with various $\gamma$ values. The results are presented in Table~\ref{tab:gamma_ablation}.
When $\gamma = 1$, the weighting follows the model’s raw attention, giving a balanced emphasis on gaze-relevant regions. For $\gamma > 1$, supervision becomes too narrow: the loss overshoots and neglects surrounding contextual cues (eyelid edges, glints, etc.), which actually hurts the gaze embedding stability. For $\gamma < 1$, it’s too diffused, blurring gaze vs. appearance signals.

\vspace{0.4pt}

\subsection{Remote Camera Experiments}
% As demonstrated, GazeShift performs well on our VR off-axis dataset. But can it generalize to the remote-camera regime?

As demonstrated, GazeShift performs well on our VR off-axis dataset. We next establish its ability to generalize to the remote-camera setting.

The core idea behind GazeShift is that the pretext task of gaze redirection can produce meaningful gaze embeddings when the primary variation between paired images is driven by changes in gaze direction. In remote camera settings, however, this assumption is less reliable due to increased variation from inconsistent eye cropping, illumination changes, and other in-the-wild factors.

To evaluate GazeShift in this regime, we aim to maximize gaze-induced variation while minimizing irrelevant appearance differences. We use two publicly available datasets representing different levels of environmental control.

\textbf{Columbia Gaze} consists of 6,000 face images from 56 subjects captured in a controlled laboratory environment. Each image is labeled with one of five discrete head poses and 21 gaze directions. To ensure that pairwise variation is gaze-specific, during training, we sample redirection pairs from within the same head pose category.

\textbf{MPIIGaze} contains 213,659 images of 15 subjects recorded in everyday conditions while seated in front of a laptop screen. Since the dataset was collected in uncontrolled environments, it exhibits significant variation in non-gaze factors such as lighting and background. To mitigate this, we sample training pairs from the same video using a short temporal window, promoting appearance similarity outside of gaze.

The input images have a resolution of 64$\times$32, in contrast to the 400×400 resolution of VRGaze. 
%Consequently, although the gaze encoder is also constructed from MobileNetV2 blocks, its architecture differs. 
Consequently, we construct the gaze encoder with a different configuration of MobileNetV2 blocks.
For appearance encoding, we employ a ResNet-18, consistent with the Cross-Encoder paper \cite{sun2021cross}.

% For the gaze encoder, we use a lightweight network built from MobileNetV2 blocks. The appearance encoder is a ResNet-18, consistent with the architecture used by Cross-Encoder.

In our first experiment, we train the model on the Columbia Gaze dataset in an unsupervised manner and evaluate its performance on MPIIGaze. Following the evaluation protocol used by Cross-Encoder, we adopt a leave-one-subject-out cross-validation scheme. Specifically, for each test subject, we train a shallow MLP regressor in a 100-shot supervised setting: 100 embeddings are randomly sampled from the remaining subjects and used as training data. This procedure is repeated for all 15 subjects in MPIIGaze, and we report the average angular error across all folds.

We compare our method to Cross-Encoder and a supervised baseline using ResNet-18 trained on three public datasets (Columbia, UTMultiview, and XGaze \cite{zhang2020eth}) as described in \cite{sun2021cross}. The results are shown in Table~\ref{table:train_on_columbia_calibrate_on_mpiigaze}. To comprehensively evaluate our architecture, we conducted two experiments with GazeShift: one equipped with our lightweight MobileNetV2-based gaze encoder and another utilizing a ResNet-18 gaze encoder. The lightweight GazeShift achieves an error of 8.00\degree, outperforming Cross-Encoder (8.32\degree) and the supervised baseline (8.35\degree) while operating with 35x fewer FLOPS and 10x fewer parameters than Cross-Encoder. Furthermore, when scaling GazeShift gaze encoder to use ResNet-18, our accuracy improves to 7.56\degree, demonstrating superior representation learning while utilizing the exact same compute budget as Cross-Encoder.

% We compare our method to Cross-Encoder and a supervised baseline using ResNet-18 trained on three public datasets (Columbia, UTMultiview, and XGaze \cite{zhang2020eth}) as described in \cite{sun2021cross}. The results are shown in Table~\ref{table:train_on_columbia_calibrate_on_mpiigaze}. GazeShift achieves an error of 8.00\degree, outperforming Cross-Encoder (8.32\degree) and the supervised baseline (8.35\degree), while using 10$\times$ fewer parameters.

In the second experiment, we follow the Cross-Encoder protocol and fine-tune the model -- initially trained on Columbia Gaze -- on MPIIGaze using unsupervised learning. For each fold, we train the model on 14 subjects and evaluate on the held-out subject using the 100-shot protocol, where a supervised MLP is trained on 100 embeddings sampled from the training subjects. Table \ref{table:mpiifinetune_Results}\ summarizes the results. The results of all models except GazeShift are taken from \cite{sun2021cross}. 
GazeShift achieves the best performance while using 10x fewer parameters.

\subsubsection{Qualitative Results on Gaze Representation}

To visually evaluate the continuity of the latent space and gaze-appearance disentanglement, we perform a latent space interpolation experiment (Figure~\ref{fig:columbia_recon}). By linearly interpolating between two target gaze vectors while conditioning on fixed appearance embeddings, the generated sequences produce a smooth, continuous eye movement. Crucially, the model preserves the unique features of the source identity without allowing appearance characteristics from the target gaze image to leak into the reconstruction. This confirms that GazeShift learns a continuous, disentangled manifold.

\begin{table}[]
\centering
\caption{Calibration results on MPIIGaze using embeddings trained on Columbia Gaze.}
\label{table:train_on_columbia_calibrate_on_mpiigaze}
\resizebox{1.0\columnwidth}{!}{%
\begin{tabular}{llccc}
\toprule
Supervision                   & Method                          & Avg. Error [$^\circ$] & Params       & FLOPS       \\ \midrule
Supervised                    & Resnet-18                       & 8.35                  & 11M          & 75M         \\ \midrule
\multirow{3}{*}{Unsupervised} & Cross-Encoder& 8.32                  & 11M          & 75M         \\
                              & GazeShift (MobileNetV2)    & 8.00         & \textbf{1M}  & \textbf{2M} \\ 
                              & GazeShift (ResNet-18) & \textbf{7.56}         & 11M & 75M\\
\bottomrule
\end{tabular}
}
\end{table}

\begin{table}[t]
    \centering
    % \small
    \caption{Unsupervised fine-tuning results on MPIIGaze.}
    \label{table:mpiifinetune_Results}
    \resizebox{0.7\columnwidth}{!}{%
    \begin{tabular}{lcc}
        \toprule
        Model & Avg. Error [$^\circ$] & Params \\
        \midrule
        ResNet-18 (ImageNet) & 10.60 & 11M \\
        Auto-encoder & 9.50 & 11M \\
        Auto-encoder (EFC) & 9.20 & 11M \\
        SimCLR & 10.00 & 25M \\
        BYOL & 11.10 & 25M \\
        Cross-Encoder & 7.20 & 11M \\
        GazeShift (MobileNetV2)& \textbf{7.15} & \textbf{1M} \\
        \bottomrule
    \end{tabular}
    }
\end{table}

\begin{figure}[t]
\includegraphics[clip, trim=1.0cm 7cm 6.0cm 0cm,width=1.0\linewidth]{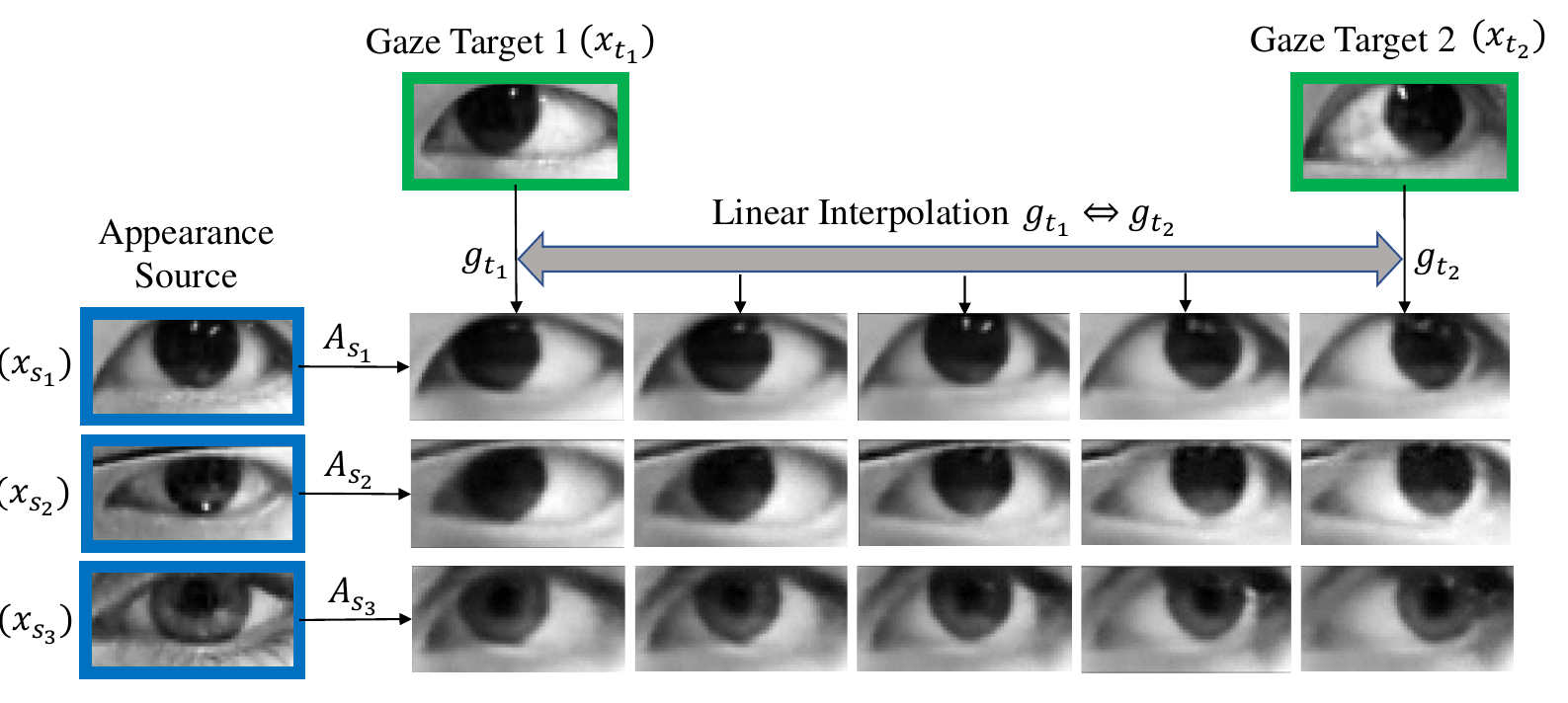}
\centering

\caption{\textbf{Latent space interpolation and gaze redirection.} We interpolate between two target gaze embeddings ($g_{t_1} \rightarrow g_{t_2}$, top green boxes) while keeping the appearance embeddings ($A_{s_1}, A_{s_2}, A_{s_3}$, left blue boxes) fixed for each row. The smooth eye movement between the gaze targets, combined with the preservation of the source images' unique appearance, visually confirms effective disentanglement and a continuous latent manifold.}

% \caption{\textbf{Latent space interpolation and gaze redirection.} We interpolate between two target gaze embeddings ($g_{t_1} \rightarrow g_{t_2}$, top green boxes) while keeping the appearance embeddings ($A_{s_1}, A_{s_2}, A_{s_3}$, left blue boxes) fixed for each row. The smooth rotation of the pupil toward the target directions, combined with the preservation of the source images' unique appearance, visually confirms effective disentanglement and a continuous latent manifold.}
% \caption{Latent space interpolation and gaze redirection. We interpolate between two distinct gaze embeddings ($g_{t1} \rightarrow g_{t2}$) extracted from target images (top green boxes). This transition is applied to three distinct appearance embeddings ($A_{s_1}, A_{s_2}, A_{s_3}$) extracted from source images (left blue boxes) and kept fixed per row. The generated sequences demonstrate a smooth, continuous rotation of the pupil that accurately reaches the target gaze directions. The preservation of each source's unique lighting and eye geometry confirms effective gaze/appearance disentanglement and a structured, continuous latent manifold.}

\label{fig:columbia_recon}
\end{figure}

\section{Limitations}
GazeShift assumes that most inter-frame appearance variation reflects gaze changes. In the VR domain, this assumption largely holds, as illumination and camera placement are tightly controlled. Appearance variations that do not correspond to gaze, such as brief blinks, eyelid motion, or pupil dilation, occur infrequently and thus have minimal effect on model performance. However, in augmented reality (AR) or mixed-reality settings, where lighting conditions and reflections are less constrained, the impact of such non-gaze appearance variations remains an open question for future investigation.

\section{Conclusion}
Despite the growing XR market, progress in VR gaze estimation remains hindered by the lack of suitable datasets and methods.
We address both challenges by introducing VRGaze, a large-scale off-axis VR dataset, and GazeShift, an unsupervised method that eliminates the need for accurate labels while achieving sub-$2^{\circ}$ error, making it suitable for practical deployment.
Implemented directly on a VR device, our gaze encoder runs in \textbf{5~ms} on the device GPU, demonstrating its value for real-time inference on edge hardware.
Beyond gaze estimation, GazeShift provides a general framework for unsupervised representation learning from paired transformations.
Its attention-guided redirection and self-supervised loss formulation can extend to other domains where appearance changes encode structured variation, such as facial motion, head-pose, or viewpoint-conditioned representations.

{
    \small
    \bibliographystyle{ieeenat_fullname}
    \bibliography{new_main}
}

%%%%%%%%%%%%%%%%%%%%%%%%%%%%%%%%%%%%%%%%%%%%%%%%%%%%%%%%%%%%%%%%%%%%

% --- Start of Supplementary Material ---
\clearpage
\appendix

% The \twocolumn[{...}] command forces the content inside to span both columns
\twocolumn[{%
\begin{center}
    \vspace*{0.5cm}
    \Large \textbf{Supplementary Material} \par
    \vspace{1.5cm}
\end{center}%
}]

% Reset counters and change numbering format to 'S'
\setcounter{section}{0}
\renewcommand{\thesection}{\Alph{section}}
\setcounter{table}{0}
\renewcommand{\thetable}{S\arabic{table}}
\setcounter{figure}{0}
\renewcommand{\thefigure}{S\arabic{figure}}
\setcounter{equation}{0}
\renewcommand{\theequation}{S\arabic{equation}}

%\begin{document}

\maketitle

\appendix

\section{GazeShift: Implementation Details and Qualitative Results}
As VRGaze and remote-camera datasets such as Columbia and MPIIGaze differ in input resolution, we adopt architecture variations tailored to each. All experiments were conducted on a single NVIDIA RTX A5000 GPU.
Our GazeShift code and VRGaze dataset are available here: \url{https://github.com/gazeshift3/gazeshift}

\subsection{VR Experimental Details}

% \begin{figure}[t]
% \includegraphics[clip, trim=1cm 4cm 19cm 0cm,width=0.9\linewidth]{figures/columbia_reconstructions.pdf}
% \centering

% \caption{Columbia Reconstructions. Source images (left column) are redirected to match the gaze direction of their corresponding target images. Targets with similar gaze directions produce similar 12D target gaze embeddings (right column), as observed in rows 2 and 3. In contrast, targets with differing gaze directions yield distinct embeddings, as shown in the bottom two rows.
% }

% \label{fig:columbia_recon}
% \end{figure}

The VRGaze input resolution is $1 \times 400 \times 400$. To support real-time inference on a VR headset, the gaze encoder is designed to be lightweight: it consists of 6 stride-2 MobileNetV2 blocks, each repeated twice and configured with a width multiplier of 2. A final linear layer projects the output to a gaze embedding of dimension $C_d$.

The appearance encoder consists of a stride-2 convolution, followed by four MobileNetV2 blocks, each repeated six times, and a final stride-1 convolutional layer, producing a feature map of size $10 \times 10 \times C_a$. The attention module employs a single-layer, single-head attention mechanism. The decoder is composed of transposed convolutional blocks, each followed by batch normalization and a \texttt{tanh} activation.

The model is trained for 20 epochs with a batch size of 30 using the AdamW optimizer, a learning rate of 0.0001, and a weight decay of 0.05. 
% \subsubsection{Qualitative comparison between GazeShift and CrossEncoder}
% For a qualitative comparison between GazeShift and Cross-Encoder, we compute final gaze predictions for each model using a per-person calibration model. We sample nine gaze targets and visualize the predictions for all subjects in the validation set. In Figure~\ref{fig:lwtl_vs_ce1}, GazeShift predictions are shown in red, Cross-Encoder in blue, and the ground-truth targets are marked with green crosses. Visually, the red points (GazeShift) align more closely with the ground truth, indicating improved accuracy.

\subsubsection{VAE baseline details}

The VAE baseline (Table 3 in the paper) employs an architecture analogous to GazeShift, utilizing a MobileNetV2-based image encoder and a matching decoder. It is trained strictly via a standard image reconstruction objective on the input frames.

\subsubsection{Supervised Appearance-based Model}

\begin{figure*}[t]
\includegraphics[clip, trim=1cm 7cm 6cm 1cm,width=1.0\linewidth]{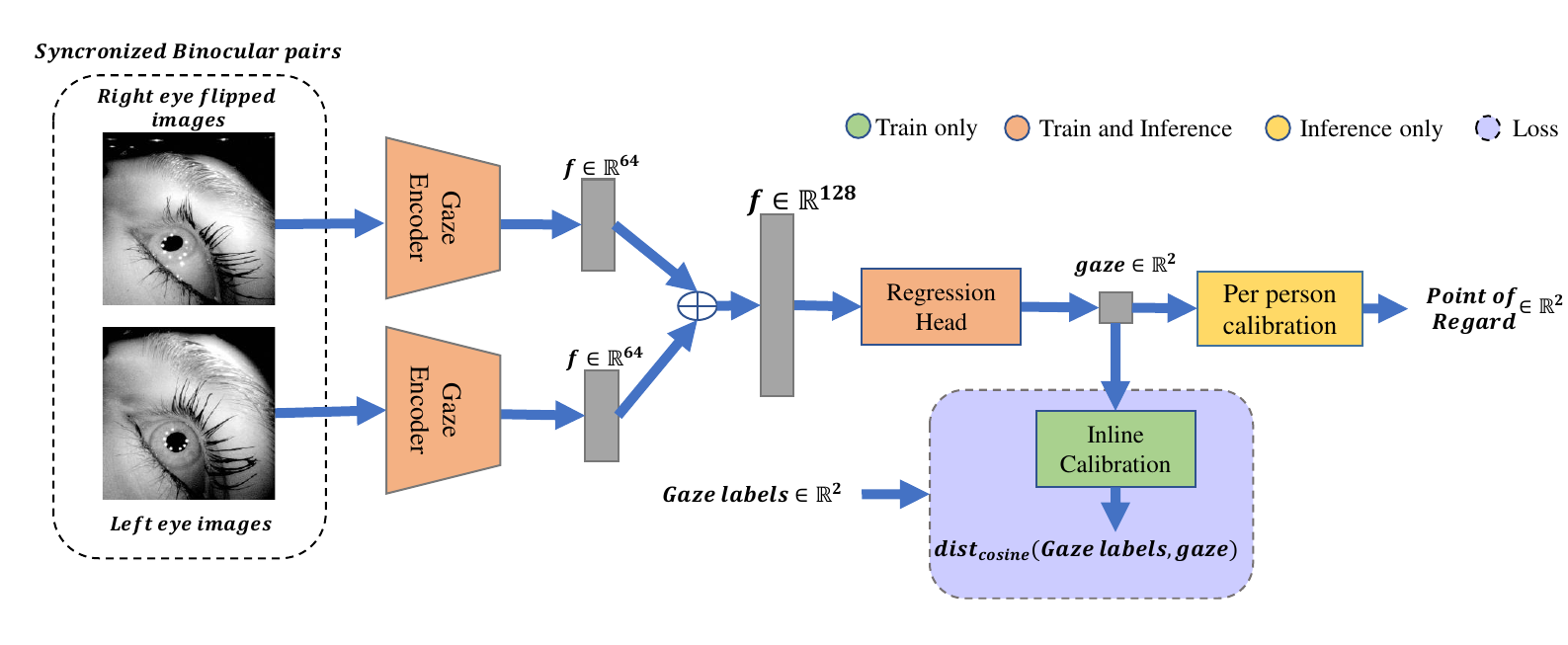}
\centering
\caption{Supervised model architecture.}

\label{fig:supervised_arch}
\end{figure*}

To contextualize the performance of GazeShift, we train a supervised binocular model using the same lightweight gaze encoder architecture. Training samples are selected from intervals where the gaze target remains stationary for three seconds, ensuring higher label reliability compared to periods of target motion. We adopt a simple Siamese configuration: synchronized left and right eye images serve as input, with the right-eye images horizontally flipped to reduce appearance asymmetry and improve representation learning. The resulting embeddings are concatenated and passed through a regression head to predict 3D gaze direction.

% The output of this regression head is passed to a per-person calibration module, which maps the raw predictions to final calibrated gaze vectors. The model is trained using a cosine distance loss between the predicted and ground-truth 3D gaze vectors.

\paragraph{Calibration aware loss.}
In conventional gaze regression models, the training loss is computed directly on raw outputs, ignoring the calibration step typically performed at inference. We propose integrating calibration directly into training by applying the loss to calibrated outputs. Specifically, we implement a linear calibration module in PyTorch that fits polynomial features to the batch-level predictions. This module is treated as a differentiable layer, allowing the model to optimize for post-calibration accuracy and better align with real-world inference behavior. Calibration aware loss reduced the average error from $1.64\degree$ to $1.54\degree$. Fig. \ref{fig:supervised_arch} depicts our supervised model architecture.

\subsection{Remote Camera Experimental Details}

In the remote camera setting, the input resolution of each eye crop is $64 \times 32$. The gaze encoder consists of 3 MobileNetV2 blocks, while the appearance encoder follows a ResNet-18 backbone, similar to the configuration used in Cross-Encoder. The attention module uses 2 layers with 2 heads, and the decoder is implemented using a DenseNet architecture.

Unlike the near-eye setting, both eye crops in the remote setup are extracted from the same image, resulting in similar appearance characteristics. We leverage this symmetry by generating source–target pairs not only from different time steps of the same eye, but also by using synchronized left/right eye pairs -- where one image is horizontally flipped to simulate a plausible gaze shift.

% \subsubsection{Qualitative Results}
% In Fig.~\ref{fig:columbia_recon}, we show eight samples from the Columbia dataset, each including a source image, target image, redirected source image, and a visualization of the target's 12D gaze embedding. The redirected source images closely resemble their corresponding targets, and similar gaze directions yield similar embeddings, highlighting the consistency of the learned gaze representation.

\section{VRGaze Dataset: Additional Characteristics and Collection Information}

%\begin{figure}[t]
%\includegraphics[clip, trim=0cm 0cm 0cm 0cm,width=0.5\linewidth]{figures/camera_position (1).png}
%\centering
%\caption{IR camera position}
%\label{fig:camera position}
%\end{figure}

\begin{figure*}[t]
\includegraphics[clip, trim=0cm 0cm 0cm 0cm,width=0.5\linewidth]{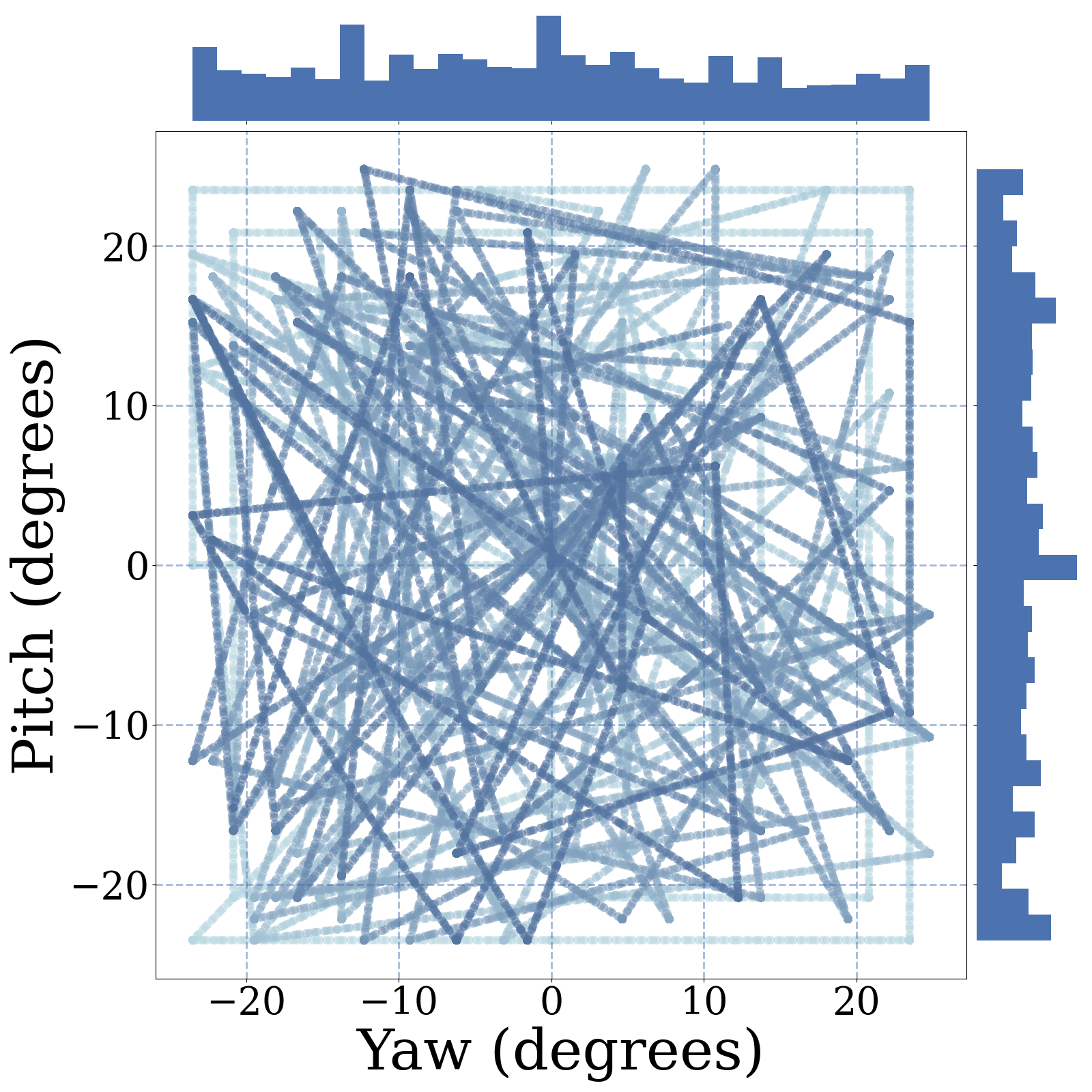}
\centering
\caption{Single person data visualization.}
\label{fig:single_person_statistics}
\end{figure*}

\paragraph{Target motion pattern.}
% \subsection{Target Motion Pattern}
Figure \ref{fig:single_person_statistics} illustrates the distribution of gaze targets for a single person over multiple sessions. Linear segments are formed by a slowly moving ball to induce smooth pursuit eye movements. The nodes at the ends of the segments are stationary points intended to elicit eye fixations. Thus, each session consists of a ball moving slowly along straight trajectories, followed by an abrupt stop lasting three seconds. During each stop, the ball gradually changes in size to enhance the fixation.  

\begin{figure*}[t]
\includegraphics[clip, trim=0cm 0cm 0cm 0cm,width=1.0\linewidth]{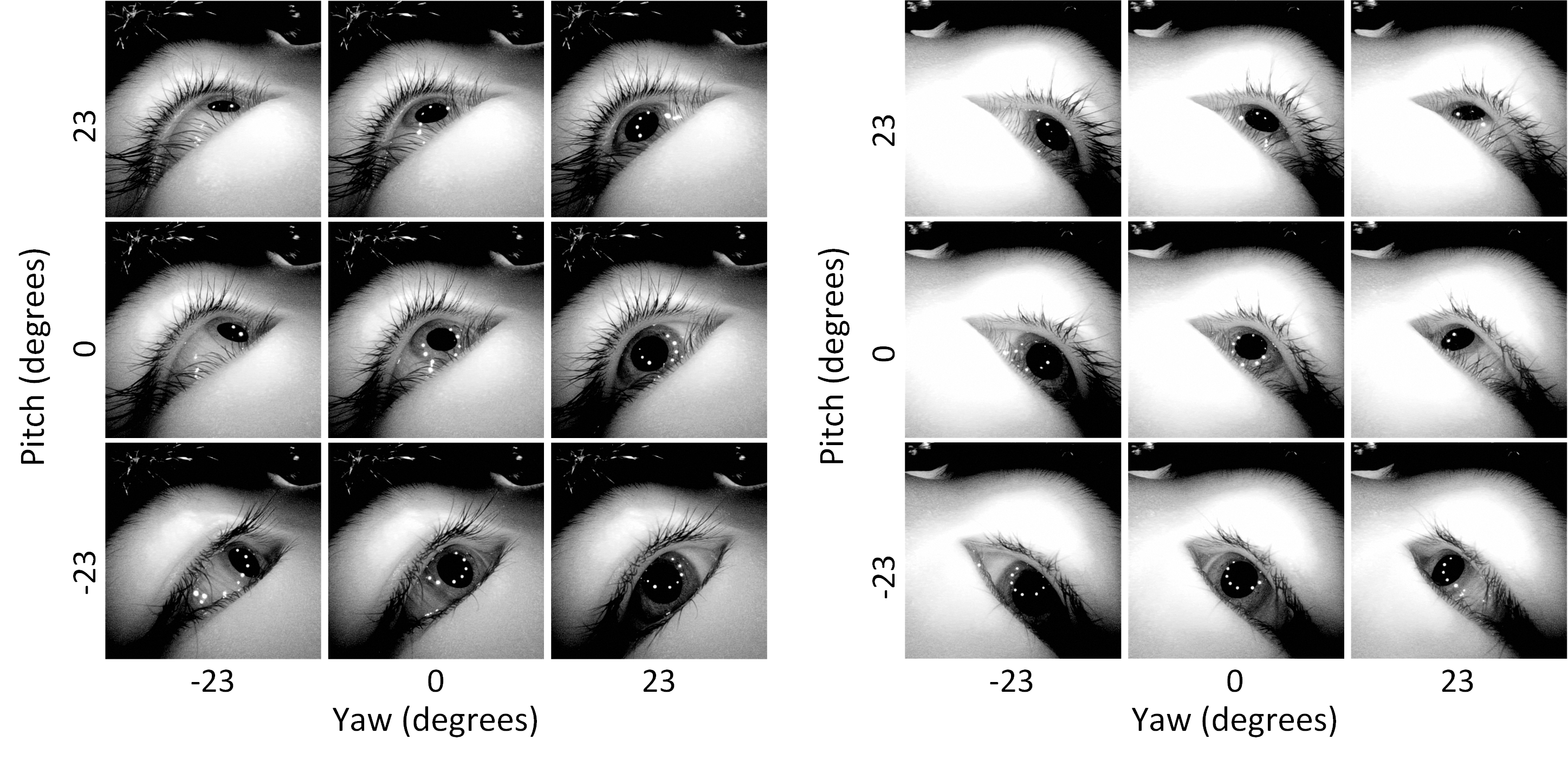}
\centering
\caption{Gaze direction visualization.}
\label{fig:gaze_direction}
\end{figure*}

\paragraph{Gaze direction diversity.}
% \subsection{Gaze Direction Diversity}
To illustrate the range of gaze directions captured, we present in Fig. \ref{fig:gaze_direction} nine example eye images corresponding to different target positions, including center, corners, and cardinal directions (left, right, up, down). This visualization demonstrates how eye appearance varies with gaze angle, providing qualitative insight into VRGaze dataset. 

\begin{figure*}[t]
\includegraphics[clip, trim=0cm 0cm 0cm 0cm,width=1.0\linewidth]{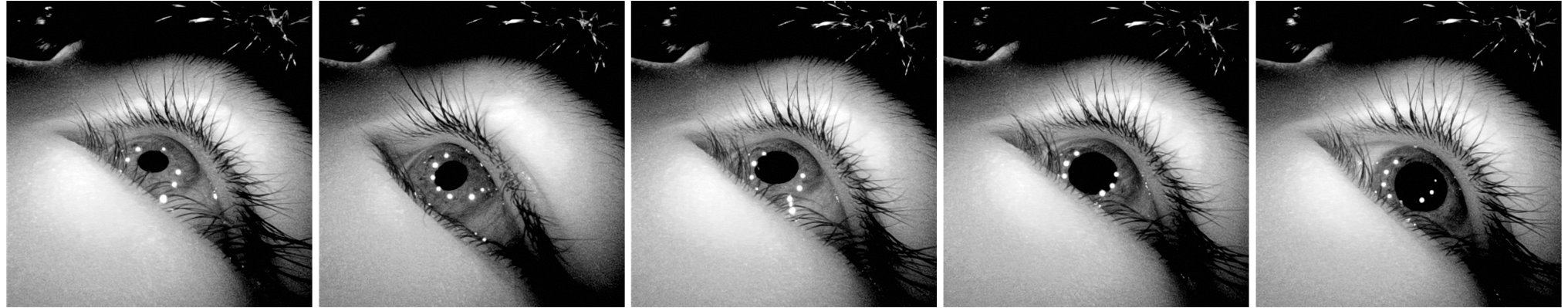}
\centering
\caption{Pupils dilation illustration.}
\label{fig:pupil_dilation}
\end{figure*}

\paragraph{Pupil dilation diversity.}
% \subsection{Pupil Dilation Diversity}
To ensure the robustness of gaze tracking under varying lighting conditions and physiological responses, we collected eye images exhibiting a wide range of pupil sizes. This was achieved by varying the background brightness during data collection sessions in VR, prompting natural pupil dilation and constriction. Fig. \ref{fig:pupil_dilation} illustrates examples of eye images with different pupil diameters, highlighting the captured diversity. Such variation is essential for training models that generalize well across users and lighting environments. 

\begin{figure*}
\includegraphics[clip, trim=0cm 0cm 0cm 0cm,width=1.0\linewidth]{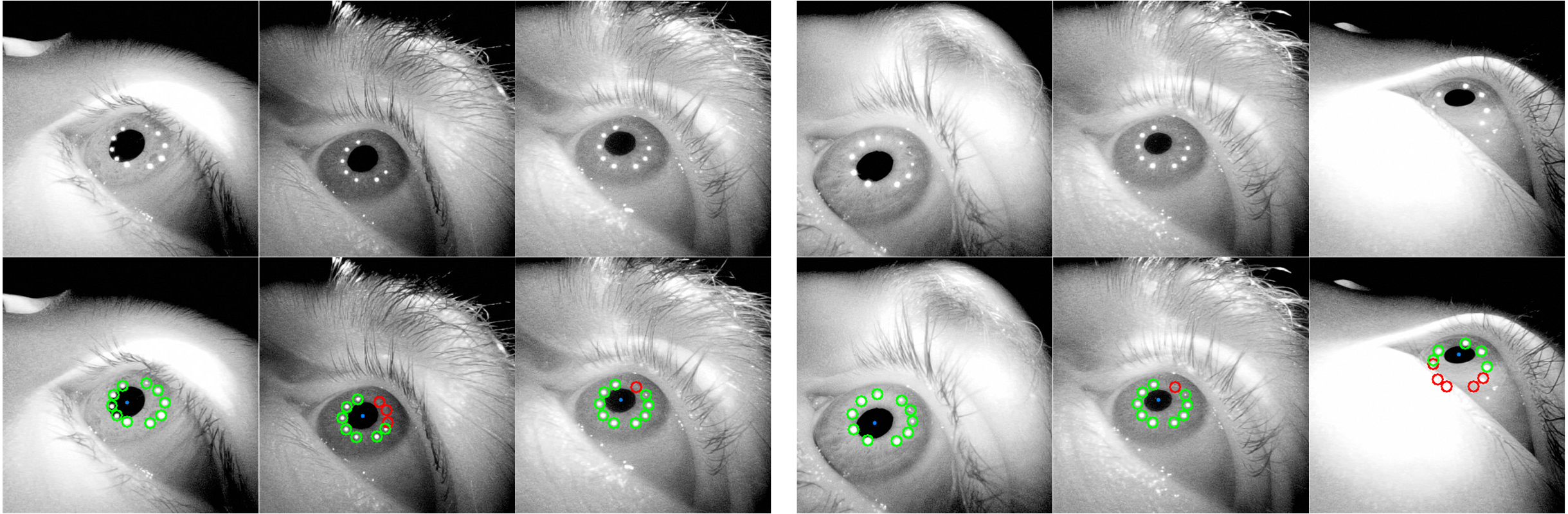}
\centering
% \caption{Glints and pupil center examples: ground truth annotations (left), predicted (right)}
\caption{Pupil center and glints from manual annotations (left) and model predictions (right).}

\label{fig:features}
\end{figure*}

\paragraph{Pupil Center and Glint Annotations}
% \subsection{Pupil center and glint annotations.}
% As part of the supplementary data to our VRGaze dataset, we also provide coordinates of the pupil center and IR LED reflections (glints) for each image. These features were not used to achieve results reported in the paper, but we have decided to include them for benefit of the research community. 
IR LED reflections (glints) are typically visible in VR cameras and may also be used for gaze estimation. To promote future research, we also release coordinates of pupil center and glints for each image in VRGaze. 
% To obtain the pupil and glints coordinates for the entire dataset, we trained a network for feature detection. We manually annotated 32k images, using 28k images for training. 
To obtain the pupil and glints coordinates for the entire dataset, we manually annotated 32k images which were used to train a feature detection model to annotate the rest. 
% The model was evaluated on a validation set of 4k images, achieving mean accuracies of 0.54 and 0.96 pixels for pupil and glint detection, respectively. 
We trained the model on 28k images and evaluated it on a validation set of 4k images, achieving mean accuracies of 0.54 and 0.96 pixels for pupil and glint detection, respectively. 
Fig. \ref{fig:features} illustrates annotated and predicted features. The pupil center is illustrated with a blue circle, visible glints with green circles, and undetected glints with red circles.

\begin{figure*}[t]
\includegraphics[clip, trim=1cm 7cm 12cm 0cm,width=0.8\linewidth]{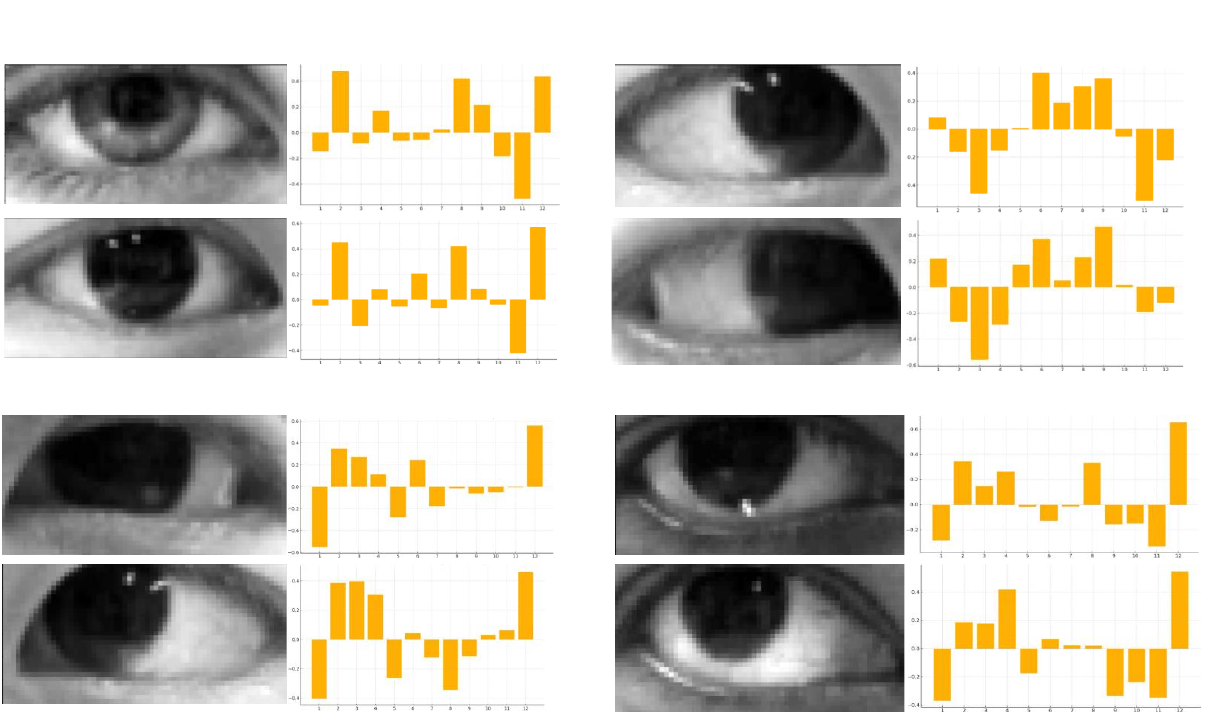}
\centering

\caption{Gaze embeddings across directions. Each quadrant shows two eye images with similar gaze directions, along with their corresponding 12D gaze embeddings. Within a quadrant, the embeddings align closely, while embeddings across quadrants differ, reflecting distinct gaze directions. Notably, in the top-left quadrant, the gaze embeddings remain similar despite clear differences in eye appearance.}

\label{fig:columbia_recon}
\end{figure*}

\begin{figure*}[t]
    \centering
    % First subfigure
    \begin{subfigure}[b]{0.48\textwidth}
        \centering
        \includegraphics[clip, trim=0cm 0cm 0cm 0.7cm,width=\textwidth]{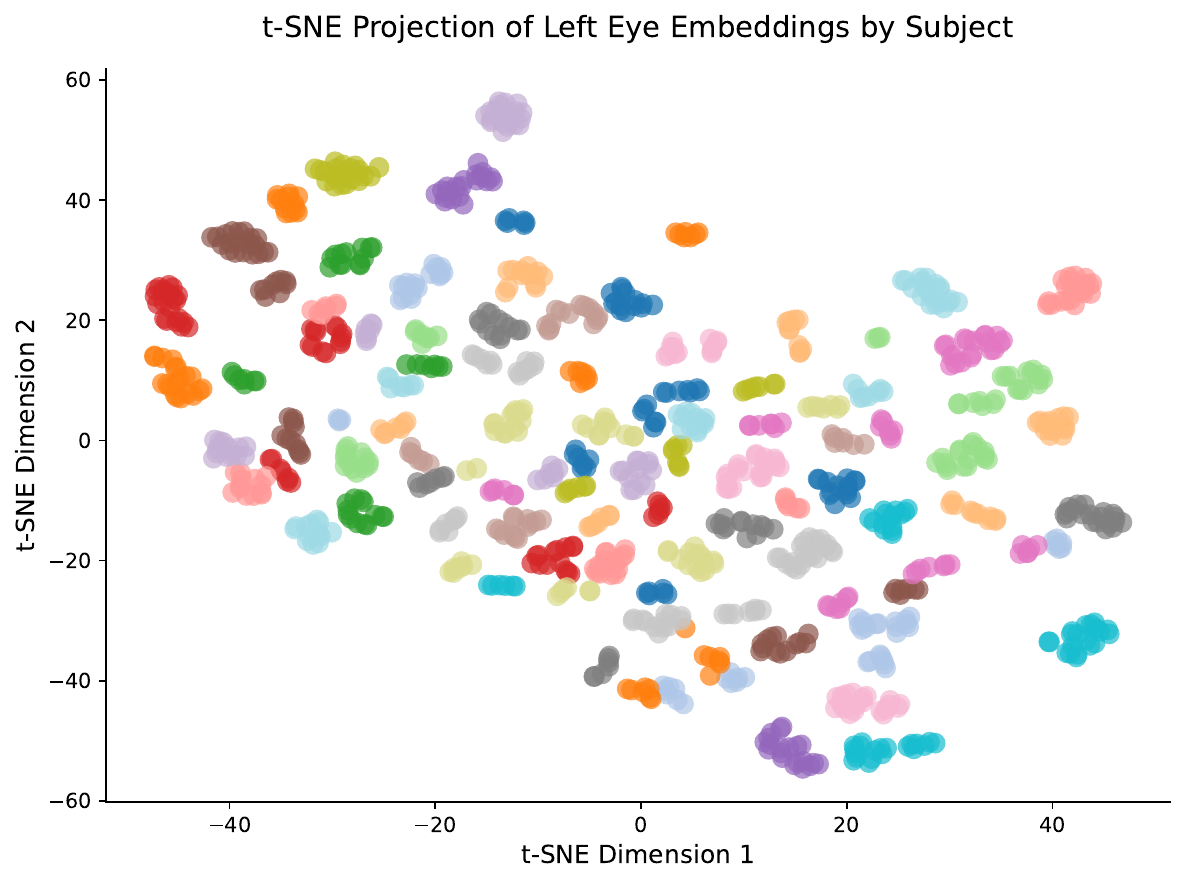}
        \caption{Appearance embeddings}
        \label{fig:sub1}
    \end{subfigure}
    \hfill % Adds spacing between the two figures
    % Second subfigure
    \begin{subfigure}[b]{0.48\textwidth}
        \centering
        \includegraphics[clip, trim=0cm 0cm 0cm 0.7cm,width=\textwidth]{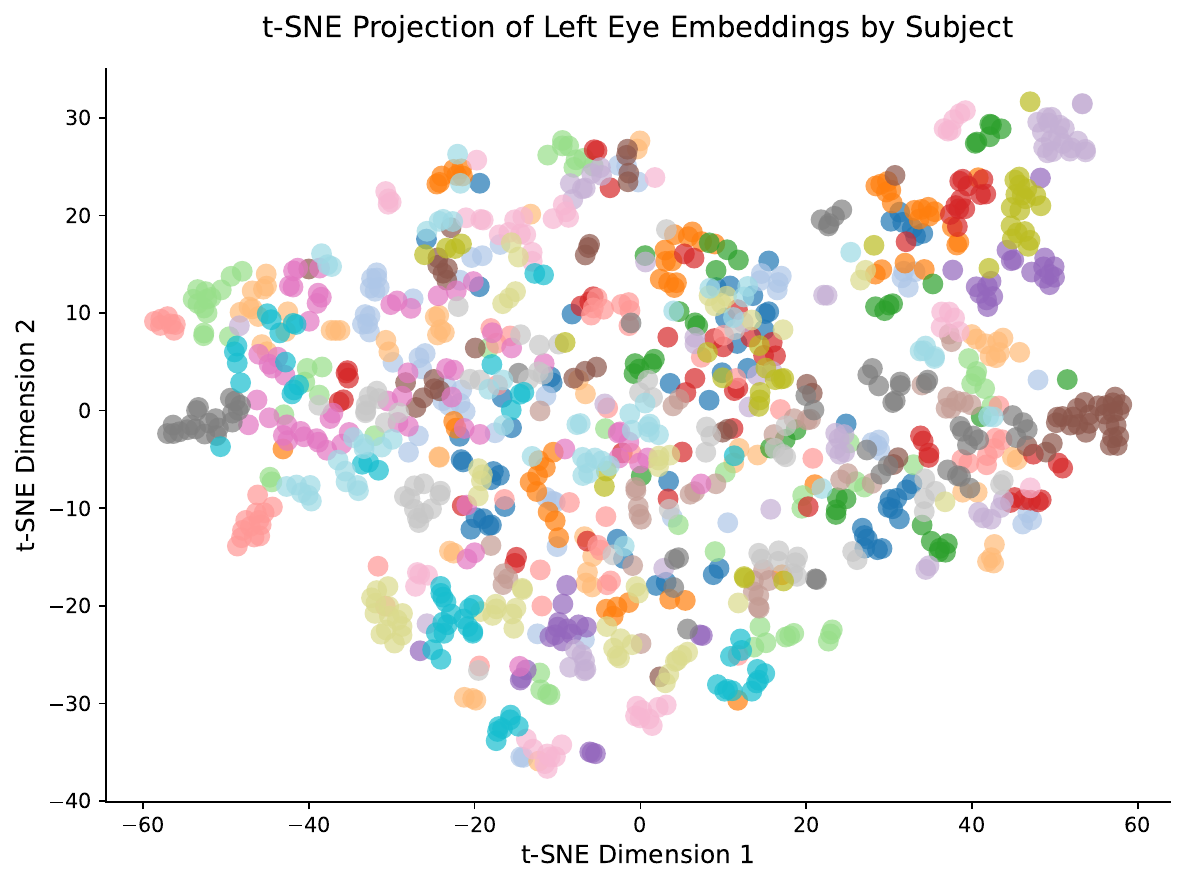}
        \caption{Gaze embeddings}
        \label{fig:sub2}
    \end{subfigure}

    \caption{\textbf{Latent space visualization.} A t-SNE projection of the representations extracted by GazeShift. Each point corresponds to an eye image from the Columbia Gaze dataset and is colored according to the subject's identity, illustrating the clustering of appearance features.}
    \label{fig:both_images}
\end{figure*}

% \begin{figure*}
% \includegraphics[clip, trim=0cm 0cm 0cm 0cm,width=1.0\linewidth]{figures/letter_of_consent.png}
% \centering
% % \caption{Glints and pupil center examples: ground truth annotations (left), predicted (right)}
% \caption{Every person participating in the VRGaze dataset signed the consent form}

% \label{fig:letter_of_consent}
% \end{figure*}

\section{VR Device specifications}

\begin{figure*}[h]
    \centering
    \includegraphics[clip, trim=1cm 2cm 6cm 0cm, width=0.8\linewidth]{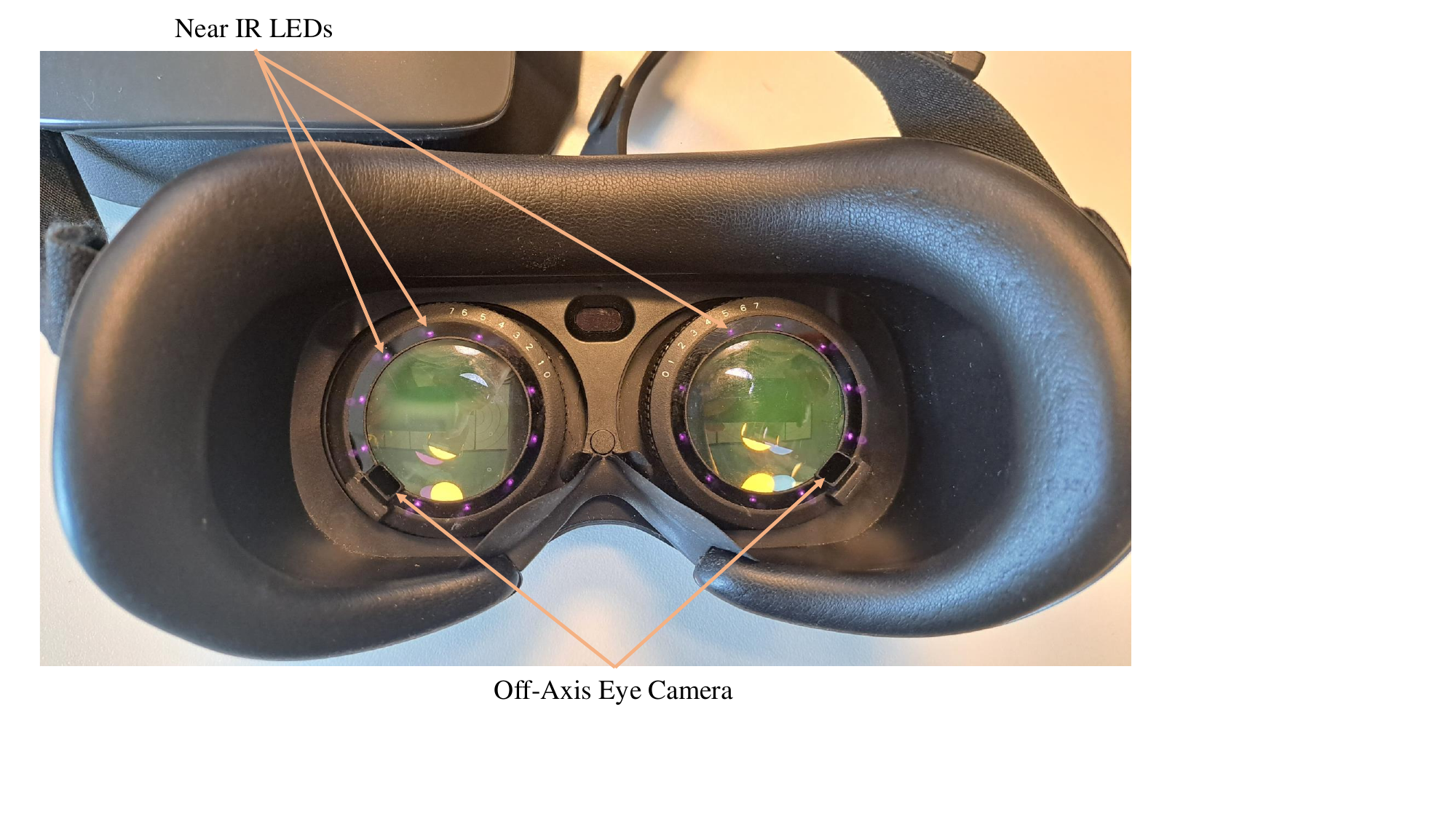}
    \caption{\textbf{VRGaze hardware setup.} The internal view of the custom virtual reality headset used for data collection. The ring of 850 nm near-infrared LEDs (visible as purple lights) provides active illumination for robust corneal glint formation, while the restrictive form factor necessitates the off-axis camera placement.}
    \label{fig:vr_hardware}
\end{figure*}

The VRGaze dataset was collected using a custom virtual reality headset prototype equipped with a dedicated off-axis eye tracking system. To validate the real-time capabilities of the GazeShift architecture, all on-device processing and inference benchmarking were executed on a Samsung Exynos 2200 System-on-Chip featuring an Xclipse 920 mobile GPU. This setup accurately reflects the thermal and computational constraints of modern standalone extended reality devices. The camera operates at 30 frames per second with a spatial resolution of 400x400 pixels, and the device provides a 90° horizontal field of view. Active illumination is supplied by an array of 10 near-infrared LEDs per eye. The camera sensor is mounted at a steep oblique angle below the eye (Figure \ref{fig:vr_hardware}) to mimic the restrictive internal layouts required by modern, slim VR form factors. Finally, strict temporal synchronization between the display rendering the visual stimulus and the camera's capture trigger ensures precise 2D Point of Regard ground truth labels across all 2.1 million recorded frames.

\end{document}